\documentclass[
twocolumn,
]{ceurart}

\usepackage{listings}

\usepackage{amsmath,amssymb,amsfonts}
\usepackage{algorithmic}
\usepackage{graphicx}
\usepackage{textcomp}
\usepackage{xcolor}
\def\BibTeX{{\rm B\kern-.05em{\sc i\kern-.025em b}\kern-.08em
    T\kern-.1667em\lower.7ex\hbox{E}\kern-.125emX}}

\usepackage{multirow}
\usepackage[]{subcaption}

\begin{document}

\copyrightyear{2024}
\copyrightclause{Copyright for this paper by its authors. Use permitted under Creative Commons License Attribution 4.0 International (CC BY 4.0).}

\conference{The IJCAI-2024 AISafety Workshop}

\title{Hyper-parameter Tuning for Adversarially Robust Models}


\author[1,2]{Pedro Mendes}[%
email=pgmendes@andrew.cmu.edu,
]
\cormark[1]
\address[1]{Software and Societal Systems Department, Carnegie Mellon University}
\address[2]{INESC-ID and Instituto Superior Tecnico, Universidade de Lisboa}

\author[2]{Paolo Romano}[%
email=romano@inesc-id.pt,
]

\author[1]{David Garlan}[%
email=dg4d@andrew.cmu.edu,
]

\cortext[1]{Corresponding author.}

\begin{abstract}
This work focuses on the problem of hyper-parameter tuning (HPT) for robust (i.e., adversarially trained) models, shedding light on the new challenges and opportunities 
arising during the HPT process for robust models.
To this end, we conduct an extensive experimental study based on three popular deep models and explore exhaustively nine (discretized)  hyper-parameters (HPs), two fidelity dimensions, and two attack bounds, for a total of 19208 configurations (corresponding to 50 thousand GPU hours).

Through this study, we show that the complexity of the HPT problem is further exacerbated in adversarial settings due to the need to independently tune the HPs used during standard and adversarial training: 
succeeding in doing so (i.e., adopting different HP settings in both  phases) can  lead to a reduction of up to 80\% and 43\% of the error for clean and adversarial inputs, respectively.
We also identify new opportunities to reduce the cost of HPT for robust models. 
Specifically, we propose to leverage cheap adversarial training methods to obtain inexpensive, yet highly correlated, estimations of the quality achievable using more robust/expensive state-of-the-art methods. We show that, by exploiting this novel idea in conjunction with a recent multi-fidelity optimizer (taKG), the efficiency of the HPT process can be enhanced by up to 2.1$\times$.

\end{abstract}


\maketitle



\section{Introduction}
Adversarial attacks~\cite{adv_train} aim at causing model misclassifications by introducing small perturbations in the input. White-box methods like Projected Gradient Descent (PGD)~\cite{pgd} have been shown to be extremely effective in synthesizing perturbations that are small enough to be noticeable by humans, while severely hindering the model's performance. Fortunately, models can be hardened against this type of attack via a so-called ``Adversarial Training'' (AT) process. During AT, which typically takes place after an initial standard training (ST) phase~\cite{delay_adv_train},  adversarial examples are synthesized and added (with their intended label) to the training set. 
Recently, several  AT methods have been proposed~\cite{adv_train,pgd,trades} that explore different trade-offs between robustness and computational efficiency. 
Unfortunately, the most robust AT methods impose significant overhead (up to $7\times$ in the models tested in this work) with respect to standard training. 

These costs are further amplified when considering another crucial phase of model building, namely hyper-parameter tuning (HPT). In fact, HPT methods require training a model multiple times using different hyper-parameter (HP) configurations. Consequently, the overheads introduced by AT lead also to an increase in the cost of HPT. 
Further, AT and ST share common HPs, which raises the question of whether AT should simply employ the same HP settings used during ST, or if a new HPT process should be executed to select different HPs for the AT phase. In the latter case, the dimensionality of the HP space to be optimized grows significantly,  exacerbating the HPT cost.

Hence, this work focuses on the problem of HPT for adversarially trained models with the twofold goal of i) shedding light on the new challenges (i.e., additional costs) that emerge when performing HPT for robust models, and ii) proposing novel techniques to reduce these costs, by exploiting opportunities that emerge in this context.

We pursue the first goal via an extensive experimental study based on 3 popular models/datasets widely used to evaluate AT methods 
(ResNet50/ImageNet, ResNet18/SVHN, and CNN/Cifar10). In this study, we discretize and exhaustively explore the HP space composed of up to nine HPs, which we evaluated considering two ``fidelity dimension''~\cite{smac,mtbo} for the training process and two attack strengths. Overall, we test a total of 19208 configurations and we make this dataset publicly accessible in the hope that it will aid the design of future HPT methods specialized for AT.

Leveraging this data, we investigate 
a key design choice for the HPT process of robust models, namely the decision of whether to adopt the same vs. different HP settings during AT and ST (for the common HPs in the 2 phases). To this end, we focus on 3 key HPs of deep models: learning rate, momentum, and batch size. Our empirical study shows that allowing the use of different HP settings during ST and AT  can bring substantial benefits in terms of model quality, by reducing up to 80\% and 43\%  the standard and adversarial error, respectively.

Further, our study demonstrates that while the cost and complexity of HPT are heightened in adversarial settings, it also reveals that, in the context of robust models,  unique opportunities can be exploited to effectively mitigate these costs.
Specifically, we show that it is possible to leverage cheap AT methods to obtain inexpensive, yet highly correlated, estimations of the quality achievable using more robust/expensive methods (PGD \cite{pgd}). Besides studying the trade-offs between cost reduction and HP quality correlation with different AT methods, we extend a recent multi-fidelity optimizer (taKG~\cite{takg}) to incorporate the choice of the AT method as an additional dimension to reduce the HPT cost. We evaluate the proposed method using our dataset and show that incorporating the choice of the AT method as an additional fidelity dimension in taKG leads up to  2.1$\times$ speed-ups, with gains that extend up to 3.7$\times$ w.r.t. popular HPT methods, as HyperBand~\cite{hyperband}.
These reductions in the optimization time not only translate to significant reductions in energy consumption during training but also result in corresponding decreases in pollutant emissions.

\section{Background and Related Work} 
\label{sec:3_related_word}

In this section, we first provide background information on AT techniques (Section~\ref{subsec:AT}) and then discuss related works in the area of HPT (Section~\ref{subsec:HPT}).

\subsection{Adversarial Training} 
\label{subsec:AT}

Adversarial attacks aim at 
introducing small perturbations to input data, often small enough to be hardly perceivable by humans, with the goal to lead the model to generate an erroneous output.
These attacks reveal vulnerabilities of current model training techniques, underscoring the need for developing robust models in different domains.
Thus, several works \cite{adv_train,pgd,fgsm_fast,fgsm_free,deepfool}  developed new techniques to mitigate these vulnerabilities and defend against adversarial attacks, by tackling them via different and often orthogonal or complementary ways, such as adversarial training~\cite{adv_train,pgd,fgsm_fast,fgsm_free},  detection of adversarial attacks~\cite{detection}, or pre-processing techniques to filter adversarial perturbations~\cite{preprocess}. 
Next, we will review existing AT approaches, which represent the focus of this work.

AT aims at improving the robustness of machine learning (ML) models by i) first generating adversarially perturbed inputs, and ii) feeding these adversarial examples, along with the correct corresponding label, during the model training phase. More formally, this process can be described as follows.
Unlike ST, which determines the models’ parameters $\theta$ by minimizing the loss function between the model's prediction for the clean input $f_{\theta }(x) $ and the original class $y$, i.e.,
$ \underset{\theta}{\min} \{  \underset{x,y \sim D}{\mathbf{E}} [L(f_{\theta}(x), y)] \}$, 
AT first computes a perturbation $\delta$, smaller than a maximum predefined bound $\epsilon$, which will mislead the current model, and then trains the model with that perturbed input. 
This approach leads to the  formulation of the following optimization problem: 
$\underset{\theta}{\min} \{ \underset{x,y \sim D}{\mathbf{E}} [ \underset{\|\delta \| < \epsilon}{\max} L(f_{\theta}(x+\delta), y) ] \}$.
The model's robustness depends on the bound $\epsilon$ used to produce the adversarial examples and on the strength of the method used to compute those examples.

Several methods have been developed to solve this optimization problem (or variants thereof as, e.g., \cite{trades}) and the resulting techniques are based on different assumptions about, e.g., the availability of the model for the attacker (i.e., white-box~\cite{pgd} or black-box~\cite{black_box_adv_train}), whether the underlying model is differentiable~\cite{adv_train} or not~\cite{black_box_adv_train},  and the existence of bounds on attacker capabilities~\cite{deepfool}.
Among these techniques, two of the most popular ones are Fast Gradient Sign Method (FGSM)~\cite{adv_train} and Projected Gradient Descent (PGD)~\cite{pgd}). Both techniques hypothesize that attackers can inject bounded perturbations and have access to a differentiable model. 
FGSM, and its later variants~\cite{fgsm_fast,fgsm_free,fgsm2}, rely on gradient descent to compute small perturbations in an efficient way. More in detail, for a given clean input $x$, this method adjusts the perturbation $\delta$ by the magnitude of the bound in the direction of the gradient of the loss function, i.e., $\delta = \epsilon \cdot \text{sign} (\nabla_{\delta} L (f_{\theta} (x+\delta), y ))$.
PGD iteratively generates adversarial examples by taking small steps in the direction of the gradient of the loss function and projecting the perturbed inputs back onto the $\epsilon$-ball around the original input, i.e., 
$\textit{Repeat:\hspace{3mm}} \delta = \mathcal{P} (\delta + \alpha \nabla_{\delta} L (f_{\theta} (x+\delta), y ) )$, where $\mathcal{P}$ is the projection onto the ball of radius $\epsilon$ and $\alpha$ can be seen as analogous to the learning rate in gradient-descent-based training. 
Due to its iterative nature, PGD incurs a notably higher computational cost than FGSM~\cite{fgsm_free,fgsm_fast}, but it is also regarded as one of the strongest methods to generate adversarial examples.
In fact, prior work has shown that PGD attacks can fool robust models trained via FGSM and that PGD-based AT produces models robust to larger perturbations~\cite{pgd}  and with higher adversarial accuracy.  
FGSM is also known to suffer from catastrophic-overfitting~\cite{catastrophic_overfitting}, in which the model's adversarial accuracy collapses after some training iterations.  
Henceforth, we will focus on FGSM and PGD, which, as mentioned, are among the most widely used and effective methods for generating adversarial examples~\cite{fgsm2}.  In fact, these methods have been extensively studied and compared in the literature~\cite{fgsm2,catastrophic_overfitting,fgsm_free,fgsm_fast,survey_at} and represent a natural starting point for investigating the trade-offs related to HPT that arise in the context of AT.

Independently of the technique used to perform AT, a relevant finding, first investigated by 
Gupta et al.~\cite{delay_adv_train}, 
is whether to use an initial ST phase before performing AT, or whether to use exclusively AT.  
This study showed that using an initial ST phase normally helps to reduce the computational cost while yielding models with comparable quality. 
This result motivates one of the key questions that we aim at answering in this work, namely whether the ST and AT phases should share the settings for their common HPs. 

\subsection{Hyper-parameter Tuning} 
\label{subsec:HPT}

HPT is a critical phase to optimize the performance of ML models. As the scale and complexity of models increase, along with the number of HP that can possibly be tuned in modern ML methods~\cite{nlp-paper-tuning}, 
HPT is a notoriously time-consuming process, whose cost can become prohibitive due to the need to repetitively train complex models on large datasets.

To address this issue, a large spectrum of the literature on HPT relies on Bayesian Optimization (BO) 
\cite{trimtuner,fabolas,ei,mtbo,takg,lynceus,smac,freeze-thawBO}. BO employs modeling techniques (e.g., Gaussian Processes) to guide the optimization process and leverages the model's knowledge and uncertainty (via a, so called, acquisition function) to select which configurations to test.
Although the use of BO can help to increase the convergence speed of the optimization process, the cost of testing  multiple HP configurations can quickly become prohibitive, especially when considering complex models trained over large datasets.

To tackle this problem, multi-fidelity techniques~\cite{smac,mtbo,fabolas,trimtuner,freeze-thawBO,takg}  exploit cheap low-fidelity evaluations (e.g., training with a fraction of the available data or using a reduced number of training epochs) and extrapolate this knowledge to recommend high-fidelity configurations. This allows for reducing the cost of testing HP configurations, while still providing useful information to guide the search for the optimal high-fidelity configuration(s)~\cite{smac,trimtuner}. 
HyperBand~\cite{hyperband} is a popular multi-fidelity and model-free approach that promotes good quality configurations to higher budgets and discards the poor quality ones using a simple, yet effective, successive halving approach~\cite{successiveHalving}. Several approaches extended HyperBand using models to identify good configurations \cite{bohb,dehb} or shortcut the number of configurations to test~\cite{hyperjump}.
While these works adopt a single budget type (e.g., training time or dataset size), other approaches, such as taKG~\cite{takg}, make joint usage of multiple budget/fidelity dimensions during the optimization process to have additional flexibility to reduce the optimization cost. taKG selects the next configuration and different budgets via model-based predictive techniques that estimate the cost incurred and information gained by sampling a given configuration for a given setting of the available fidelity dimensions.

In the area of HPT for robust models, the work that is more closely related to ours is the study by 
Duesterwald et al.~\cite{hp_robust_ibm}. 
This work has investigated empirically the relations between the bounds on adversarial perturbations ($\epsilon$) and the model's accuracy/robustness. Further, it showed that the ratio of clean/adversarial examples included in a batch (during AT) can have a positive impact on the model's quality and represents, as such, a key HP. Based on this finding, we incorporate this HP among the ones tested in our study. 
Differently from that work, we focus on 
i) quantifying the benefits of using different HPs during ST and AT, 
and ii) exploiting the correlation between cheaper AT methods (such as FGSM) to enhance the efficiency of multi-fidelity HPT algorithms.

\section{HPT for Robust Models: Challenges and Opportunities} 
\label{sec:study}

As mentioned, this work aims at shedding light on the challenges and opportunities that arise when performing HPT for adversarially robust models.
More precisely, we seek to answer the following questions:
\begin{enumerate}
  \item Should the HPs that are common to the AT and ST phases be tuned independently? More in detail, we aim at quantifying to what extent the model's quality is affected if one uses the same vs different HP settings during the ST and AT phases 
  (see Section~\ref{sec:6_HPO}).\\

  \item Is it possible to reduce the cost of HPT by testing HP settings using cheaper (but less robust) AT methods? How correlated are the performance of alternative AT approaches and what factors (e.g., the perturbation bound or the cost of the techniques) impact such correlation? To what extent can this approach enhance the HPT's process efficiency?  (see Section~\ref{sec:7_multiBudgets}.)
\end{enumerate}

In order to answer the above questions, we have collected (and made publicly available) a dataset, which we obtained by varying some of the most impactful HPs for three popular neural models/datasets and measured the resulting model quality. We provide a detailed description of the dataset in Section~\ref{sec:experiment}.

\subsection{Experimental Setup}
\label{sec:experiment}

We base our study on three widely-used models and datasets (ResNet50/ImageNet, ResNet18/SVHN, and  CNN/Cifar10). 
All the models were trained using 1 worker, except SVHN, in which two workers were used.
We used Nvidia Tesla V100 GPUs to train the ResNet50, and  Nvidia  GeForce RTX 2080 for the remaining models. All models and training procedures were implemented in Python3 via the Pytorch framework.

To evaluate the models, we considered up to nine different HPs, as summarized in Table~\ref{tab:fixed_hp}. The first three HPs in this table apply to both the ST and AT phases.  $\alpha$ is an HP that applies exclusively to AT, whereas the last two HPs (\%RAT and \%AE) regulate the balance between ST and AT (see Section~\ref{sec:3_related_word}). Specifically, \%RAT defines the number of computational resources allocated to the AT phase, and \%AE indicates the ratio of adversarial inputs contained in the batches during the AT phase (as suggested by 
Duesterwald et al.~\cite{hp_robust_ibm}).
We further consider several settings of the bound $\epsilon$ on the attacker power (see Table~\ref{tab:bounds}).
Note that the reported values of $\epsilon$ are normalized by 255. 
Finally, we also consider two fidelity dimensions, namely the number of training epochs and the number of PGD iterations (see Section~\ref{sec:7_multiBudgets}). The model's quality is evaluated using standard error (i.e., error using clean inputs) and adversarial error (i.e., error using adversarially perturbed inputs).

For each model, we exhaustively explored the (discretized) space defined by the HPs, bound $\epsilon$, and fidelities, which yields a search space encompassing a total of 19208 configurations. Building this dataset required around fifty thousand GPU hours and we have made it publicly accessible in the hope that it will aid the design of future HPT methods specialized for AT. Additional information to ensure the reproducibility of results is provided in the public repository\footnote{\url{https://github.com/pedrogbmendes/HPT_advTrain}}.

\begin{table}[t]
\centering
    \caption{Hyper-parameters considered}
    \label{tab:fixed_hp}
    \begin{tabular}{cc} 
    \toprule
    \textbf{Hyper-parameter} & \textbf{Values} \\
    \midrule
    Learning Rate (ST and AT)  &  \{0.1, 0.01\}  \\\hline 
    \multirow{2}{*}{Batch Momentum (ST and AT)}   & \{0.9, 0.99\} for ResNet50  \\
    &  \{0, 0.9\} otherwise\\\hline 
    \multirow{2}{*}{Batch Size (ST and AT)}  & \{256, 512\} for ResNet50 \\
    &  \{128, 256\} otherwise\\\hline

     $\alpha$ (PGD learning rate)& \{$10^{-2}$, $10^{-3}$\} \\ \hline 
     %
     %
     \% resources (time or epochs) & \multirow{2}{*}{ \{0, 30, 50, 70, 100\}} \\
    for AT (\%RAT) &  \\\hline
      %
    \% adversarial examples  & \multirow{2}{*}{ \{30, 50, 70, 100\} }\\
    in each batch (\%AE) & \\
    \bottomrule
    \end{tabular}
\vspace{2mm}

\centering
    \caption{Bounds $\epsilon$ per benchmarks}
    \label{tab:bounds}
    \begin{tabular}{ cc } 
    \toprule
        \textbf{Model \& Benchmark} & \textbf{Bound $\epsilon$} \\
    \midrule
        ResNet50/ImageNet & \{2, 4\} \\
        ResNet18/SVHN & \{4, 8\} \\
        CNN/Cifar10 & \{8, 12\} \\
    \bottomrule
    \end{tabular}
\vspace{2mm}


\centering
    \caption{Fidelities considered}
    \label{tab:budgets}
    \begin{tabular}{ cc } 
    \toprule
        \textbf{Fidelities}  &\textbf{Values} \\
    \midrule
        PGD iterations &  \{1 (FGSM), 5, 10, 20\} \\
        Epochs & \{1,2,4,8,16\} \\
    \bottomrule
        \end{tabular}

\end{table}

\subsection{Should the HPs of ST and AT be tuned independently?} 
\label{sec:6_HPO}

This section aims at answering the following question: 
given that the ST and AT phases share several HPs (e.g., batch size, learning rate, and momentum in the models considered in this study), how relevant is it to use different settings for these HPs in the two training phases? 
Note that, if we assume the existence of $C$ HPs in common between ST and AT, then enabling the use of different values for these HPs in each training stage causes a growth of the dimensionality of the HP space from $C$ to $2C$ (not accounting for any HP not in common) and, ultimately, to a significant increase in the cost/complexity of the HPT problem. Specifically, for the scenarios considered in this study, the cardinality of the HP space grows from 320 to 2560 distinct configurations. Hence, we argue that such a cost is practically justified only if it is counterbalanced by relevant gains in terms of error reduction.
To answer this question, we trained the models during 16 epochs and used different settings for the common HPs for ST and AT (Table~\ref{tab:fixed_hp}).
We consider three different settings (30\%, 50\%, and 70\%)\footnote[2]{We exclude the cases \%RAT=\{0,100\} in this study to focus on scenarios that contain both the ST and AT phases.} for the relative amount of resources (epochs) available for AT (\%RAT), as well as different settings of the perturbation bound $\epsilon$.

\begin{figure*}[t]
\centering
    \begin{subfigure}[b]{0.6\textwidth}
        \includegraphics[width=\textwidth]{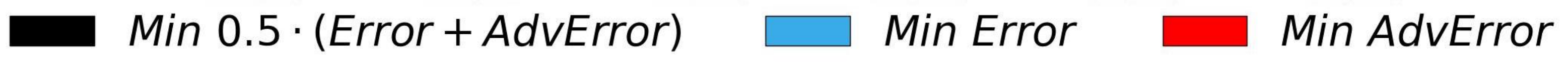}
    \end{subfigure}

    \begin{subfigure}[b]{0.32\textwidth}
        \includegraphics[width=\textwidth]{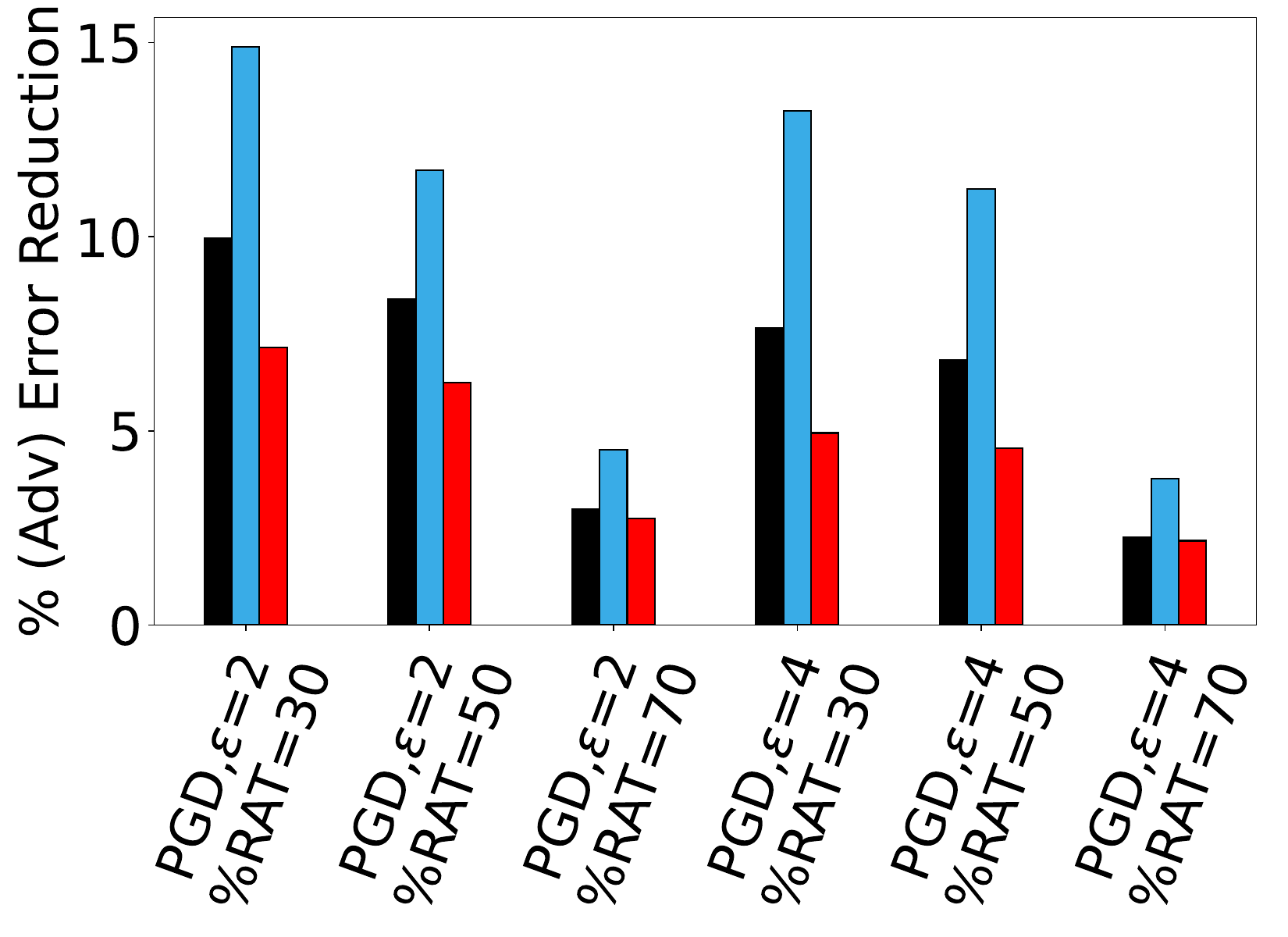}
        \caption{ResNet50/ImageNet}
        \label{fig:acc_red_imagenet}
    \end{subfigure}
    \begin{subfigure}[b]{0.32\textwidth}
        \includegraphics[width=\textwidth]{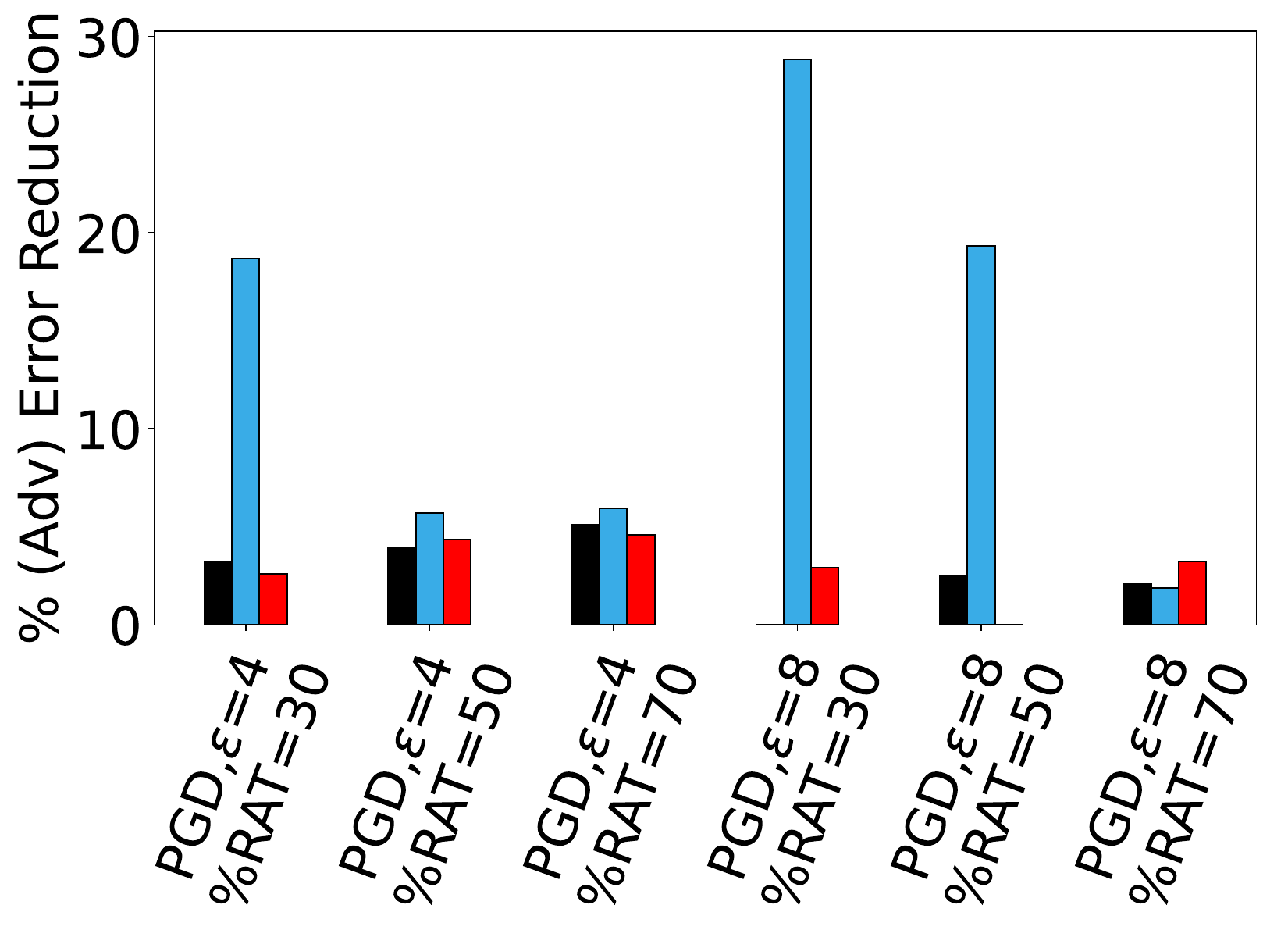}
        \caption{ResNet18/SVHN}
        \label{fig:acc_red_svhn}
    \end{subfigure}
    \begin{subfigure}[b]{0.32\textwidth}
        \includegraphics[width=\textwidth]{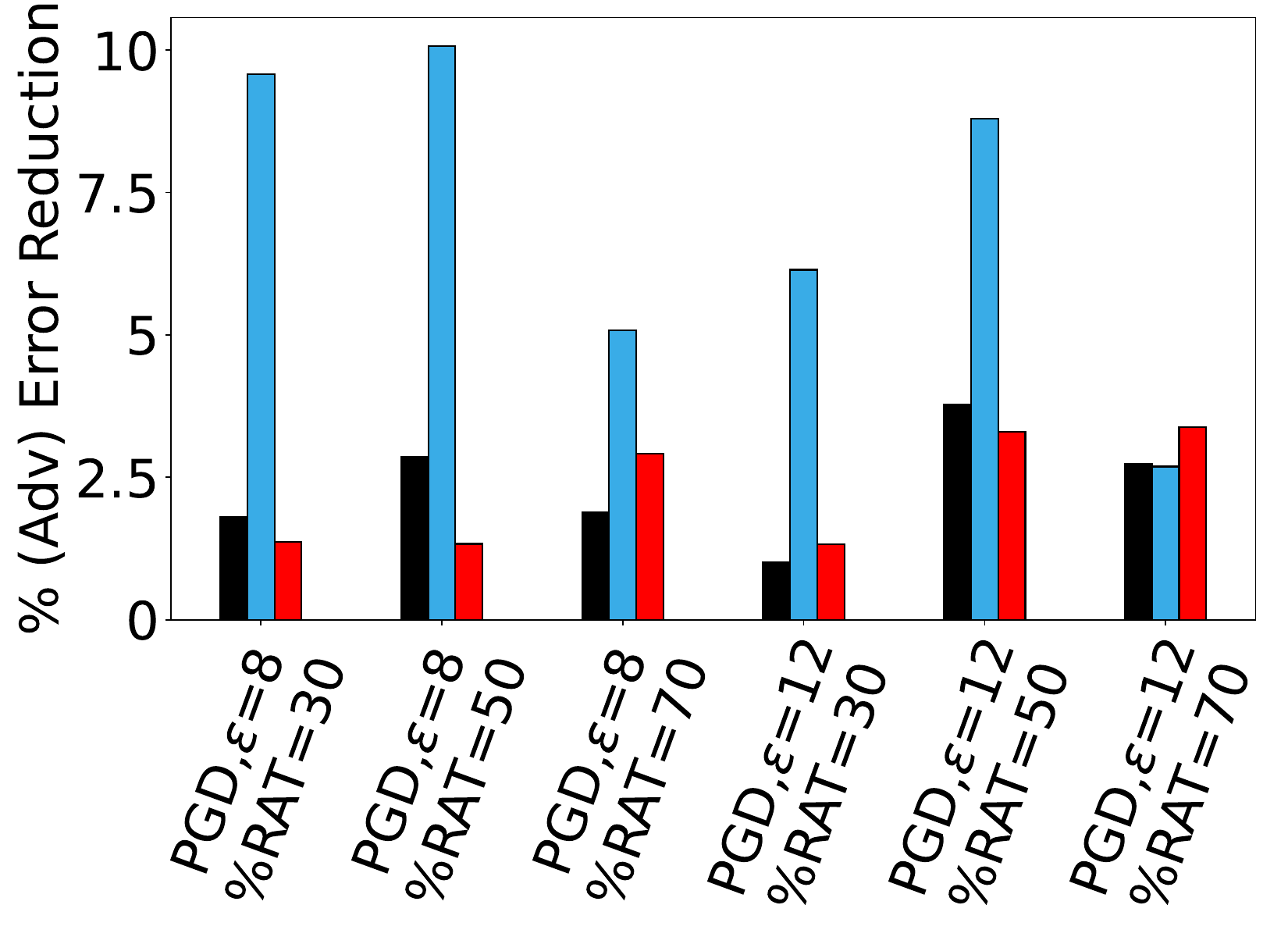}
        \caption{CNN/Cifar10}
        \label{fig:acc_red_cifar10}
    \end{subfigure}

    \caption{Reduction of the mean (in black), standard (in blue), and adversarial (in red)  error of the optimal configuration if the same or different hyper-parameters are used for the 2 phases of training using different scenarios and benchmarks.}
     \label{fig:acc_red}



\vspace{3mm}
    \begin{subfigure}[b]{0.32\textwidth}
    \centering
        \includegraphics[width=\textwidth]{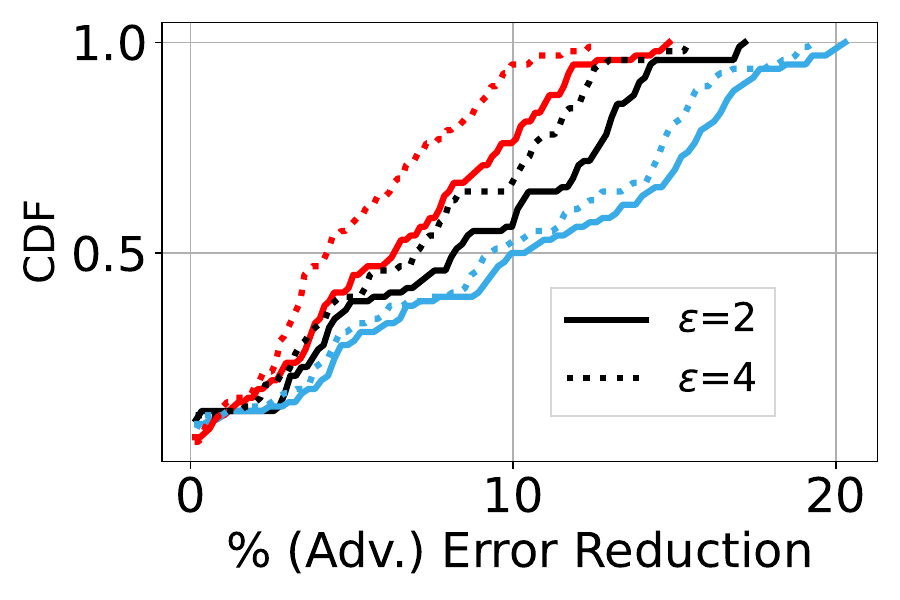}
        \caption{ResNet50/ImageNet}
        \label{fig:cdf_red_imageNet_pgd4}
    \end{subfigure}
    \begin{subfigure}[b]{0.32\textwidth}
    \centering
        \includegraphics[width=\textwidth]{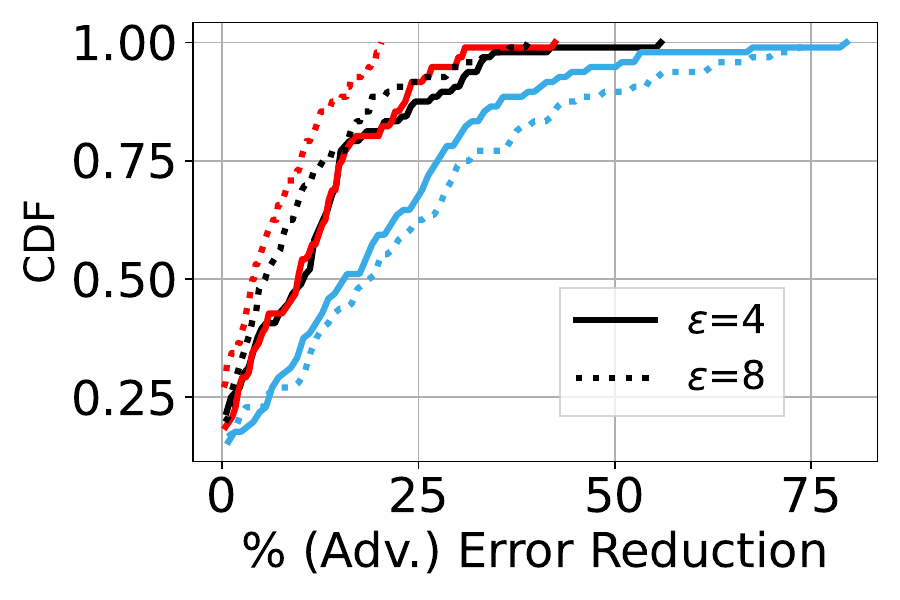}
        \caption{ResNet18/SVHN}
        \label{fig:cdf_red_svhn_pgd8}
    \end{subfigure}
    \begin{subfigure}[b]{0.32\textwidth}
    \centering
        \includegraphics[width=\textwidth]{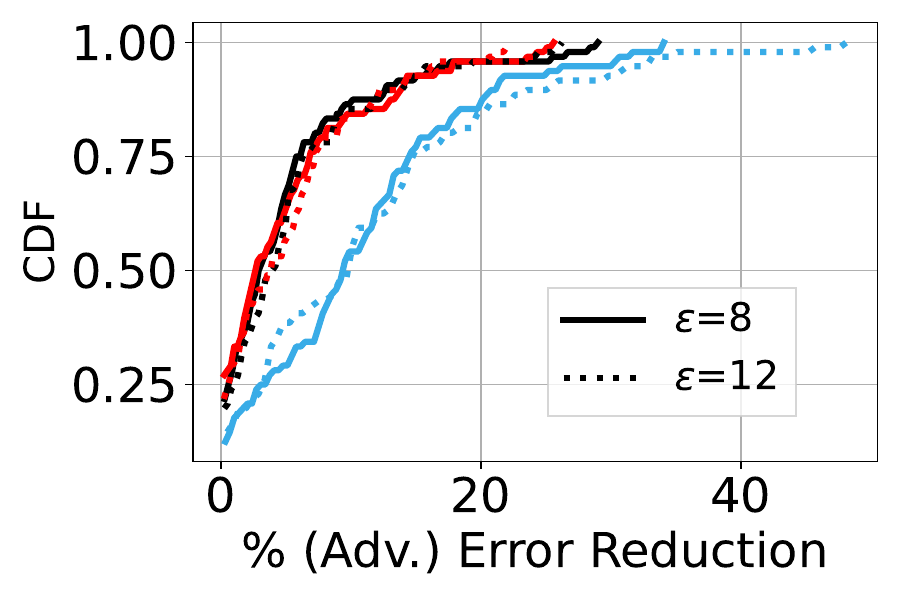}
        \caption{CNN/Cifar10}
        \label{fig:cdf_red_cifar10_pgd12}
    \end{subfigure}
    
    \caption{Cumulative distribution functions (CDFs) of the mean (in black),  standard (in blue), and adversarial  (in red) error reduction when using different HP settings for ST and AT w.r.t. the case in which common HPs are used in both phases. We use dashed and continuous lines to refer to different values of $\epsilon$.}
     \label{fig:cdf_acc_red1}
\end{figure*}

We consider that the model's HPs can be optimized according to three criteria: i) clean data error (Error), ii) adversarial error (AdvError), and iii) the average of clean and adversarial error (MeanError). For each  of these three optimization criteria, \%RAT and bound $\epsilon$,  we report in Figure~\ref{fig:acc_red} the percentage of reduction of the target optimization metric for the optimal HP configuration obtained by allowing for (but not imposing) different settings of the  HPs in common to the ST and AT phases, with respect to the optimal HP configuration if one opts for using the same settings for the common HPs in both training phases, namely: 

\begin{equation}
    \text{\%Error Reduction} = 
    \frac{\text{Error}_{\text{same  HPs} } - \text{Error}_{\text{diff  HPs}}}{\text{Error}_{\text{same  HPs}}} \times 100 
\end{equation}

The results show that adopting different HP settings in the two phases can lead to significant error reductions for all the three optimization criteria. The peak gains extend up to approx. 30\% and are achieved for the case of ResNet18 with (relatively) large values of $\epsilon$ and when allocating a low percentage of  epochs  to AT (\%RAT=30\%). Overall,  the geometric means of the \% error reduction (across all models and settings of $\epsilon$ and \%RAT) is 9\%, 5\%, and 6\% for the Error, AdvError, and  MeanError criterion, respectively.




Next, Figure~\ref{fig:cdf_acc_red1} provides a different perspective in order to quantify the benefits achieved by 
separately optimizing the HP of the AT phase (vs. using for the AT phase the same HPs settings in common with the ST phase), assuming to have executed an initial ST phase using any of the possible HPs settings. Specifically, we report the Cumulative Distribution Function (CDF) of the percentage of error reduction (for each of the three target optimization metrics), when allowing the use of the same or different common HP settings for the two phases and while varying the remaining (non-common) HPs, namely \%RAT, \%AE, and $\alpha$. 
These results allow us to highlight that by independently tuning the HPs of the two training stages, the model's quality is enhanced by up to approx. 80\%, 43\%, and 56\%, when minimizing the standard,  adversarial, or mean error, resp.

Overall, these results may be justified by considering that the optimization objectives and constraints of the ST and AT phases are different, hence benefiting from using different HP settings. During ST, the training procedure focuses on maximizing standard accuracy, and the model's goal is to learn representations that generalize well to new data. 
In contrast, AT seeks to increase robustness against adversarial attacks, and the model needs to learn to differentiate between clean and perturbed examples correctly. Further, the AT phase benefits from a pre-trained model (using clean data), and, as such, this model is expected to require relatively small weight adjustments to defend against adversarial inputs. Thus, this phase is likely to benefit from more conservative settings of HPs such as learning rate and momentum than the initial ST, whose convergence could be accelerated via the use of more aggressive settings for the same HPs.
In fact, we confirmed this fact by analyzing the configurations that yield the 10 largest error reductions in Figure~\ref{fig:cdf_acc_red1}: better quality models used lower learning rates and batch sizes in the AT phase.

Another factor that can justify the need for using different HP settings during ST and AT is related to the observation that the bound on the admissible perturbation ($\epsilon$) can have a deep impact on the model's performance, by exposing an inherent (and well-known~\cite{robustness}) trade-off: as the bound increases, the model may become more robust to adversarial inputs but at the cost of an increase in the misclassification rate of clean inputs. To achieve an optimal trade-off between robustness and accuracy, it may be necessary to adjust the tuning of the HPs used during AT as $\epsilon$ varies, which in turn implies that the optimal HPs settings used during ST and AT can be different. 
In fact, by analyzing the results obtained on ResNet18/SVHN, for example, we see that the amplitude of the bound has an impact on the (adversarial) error reduction achievable by 
tuning independently the HPs of two phases of training: the 90$^{th}$ percentile of the percentage of clean error reduction is 50\% and 65\% using $\epsilon$=4 and $\epsilon$=8, respectively (see Fig.~\ref{fig:cdf_red_svhn_pgd8}).

\subsection{Can cheap AT methods be leveraged to accelerate HPT?}
\label{sec:7_multiBudgets}

\begin{figure*}[t]
\centering
    \captionsetup{justification=centering,}
    \begin{subfigure}[b]{0.24\textwidth}
        \includegraphics[ width=\textwidth]{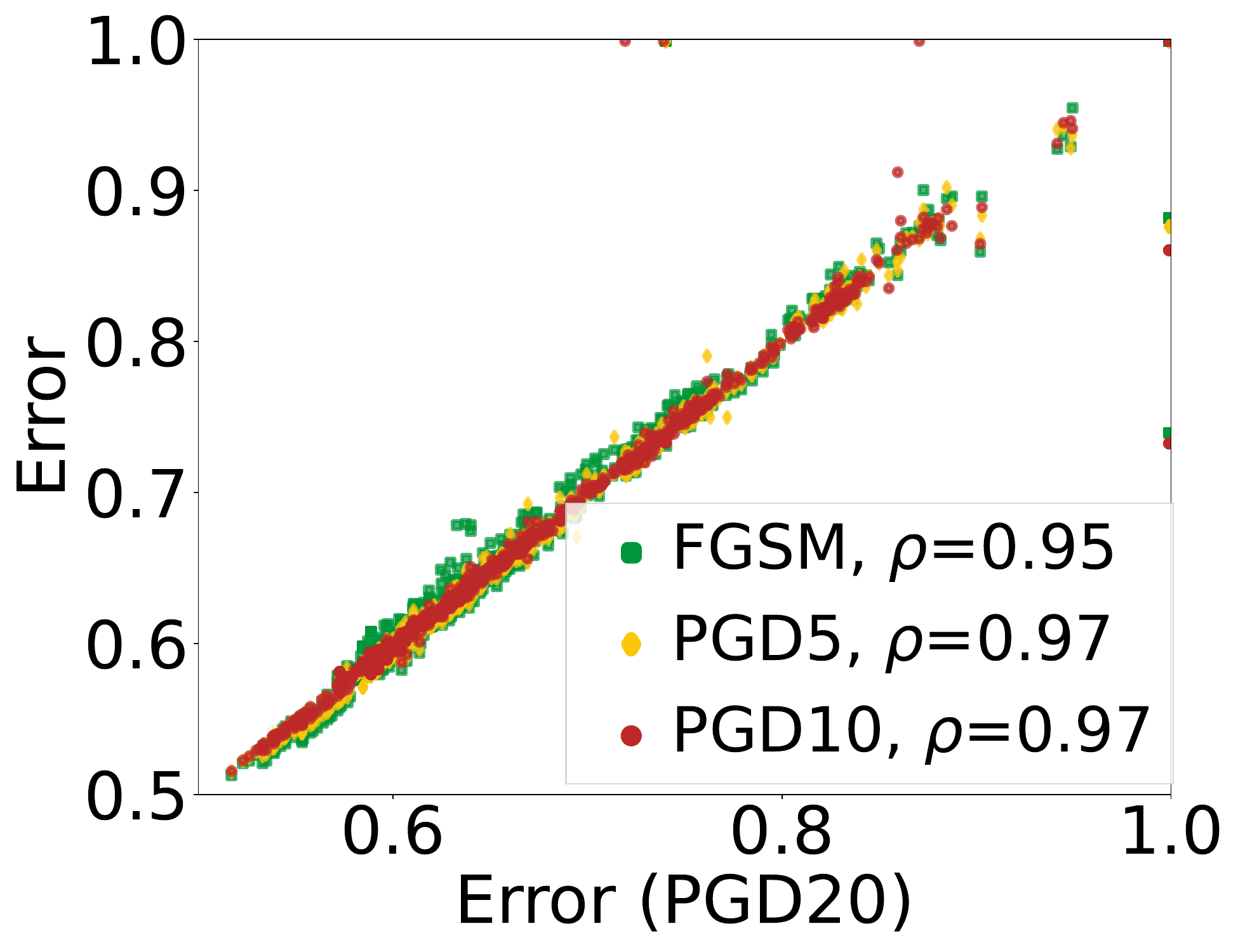}
        \caption{Error Corr. \\ResNet50/ImageNet $\epsilon$\textdblhyphen2}
        \label{fig:correlation_acc_imageNet_2}
    \end{subfigure}\hspace{2mm}
    \begin{subfigure}[b]{0.24\textwidth}
        \includegraphics[width=\textwidth]{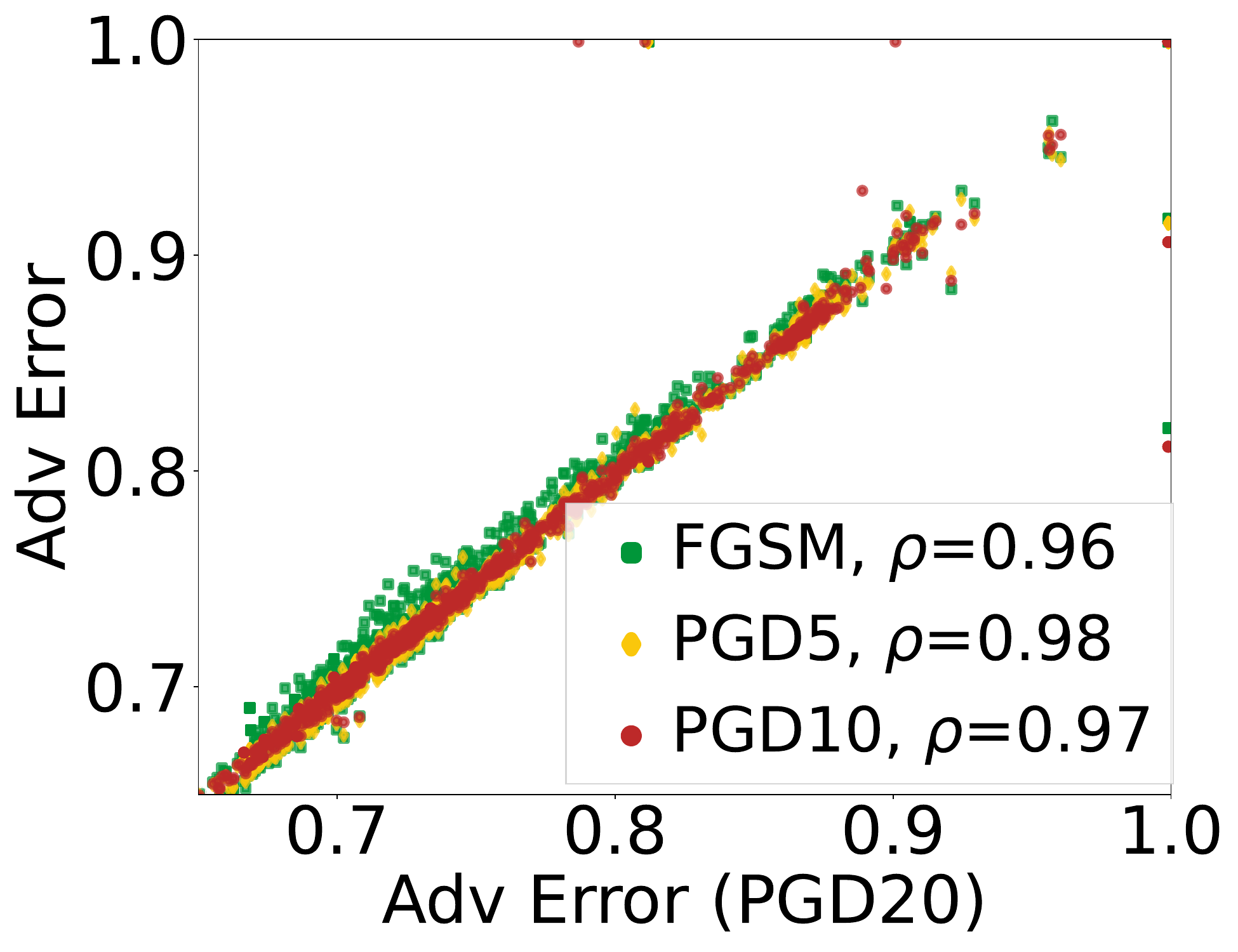}
        \caption{Adv. Error Corr. \\ResNet50/ImageNet $\epsilon$\textdblhyphen2}
        \label{fig:correlation_advacc_imageNet_2}
    \end{subfigure}\hspace{2mm}    
    \begin{subfigure}[b]{0.24\textwidth}
        \includegraphics[ width=\textwidth]{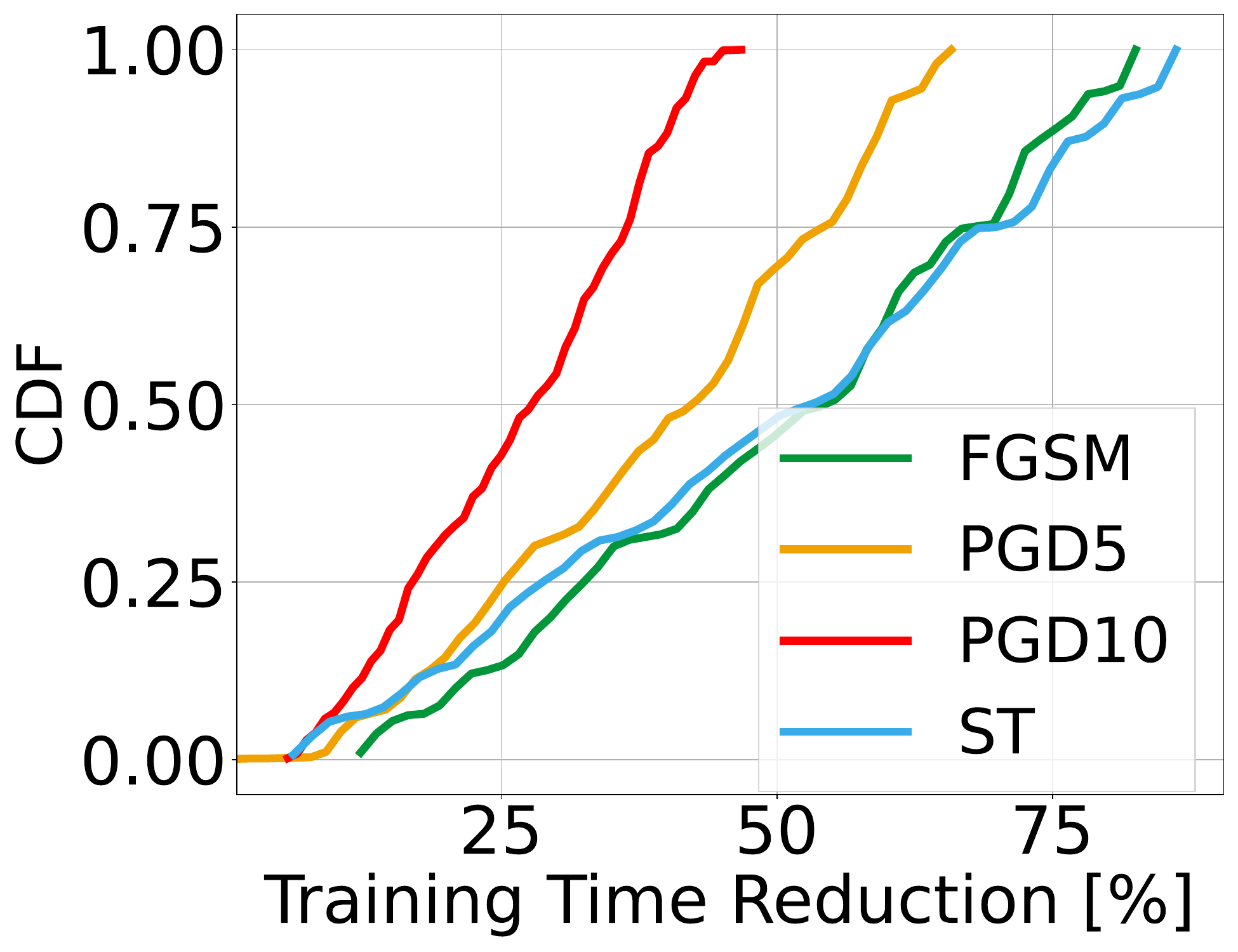}
        \caption{CDF time reduction  \\ResNet50/ImageNet}
        \label{fig:cost_imageNet}
    \end{subfigure}

\vspace{3mm}
    \begin{subfigure}[b]{0.24\textwidth}
        \includegraphics[ width=\textwidth]{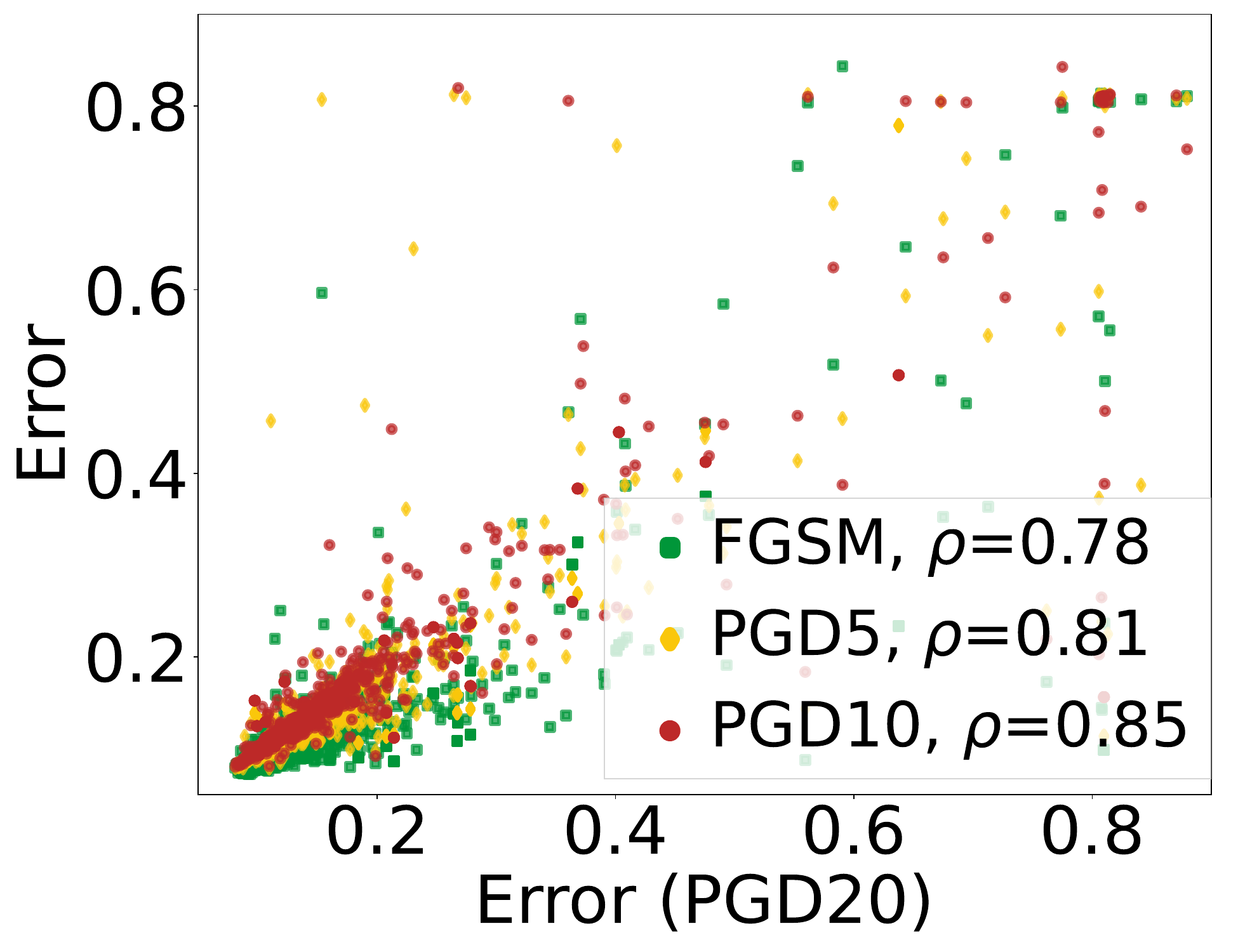}
        \caption{Error Corr. \\ResNet18/SVHN $\epsilon$\textdblhyphen4}
        \label{fig:correlation_acc_svhn_4}
    \end{subfigure}
    \begin{subfigure}[b]{0.24\textwidth}
        \includegraphics[width=\textwidth]{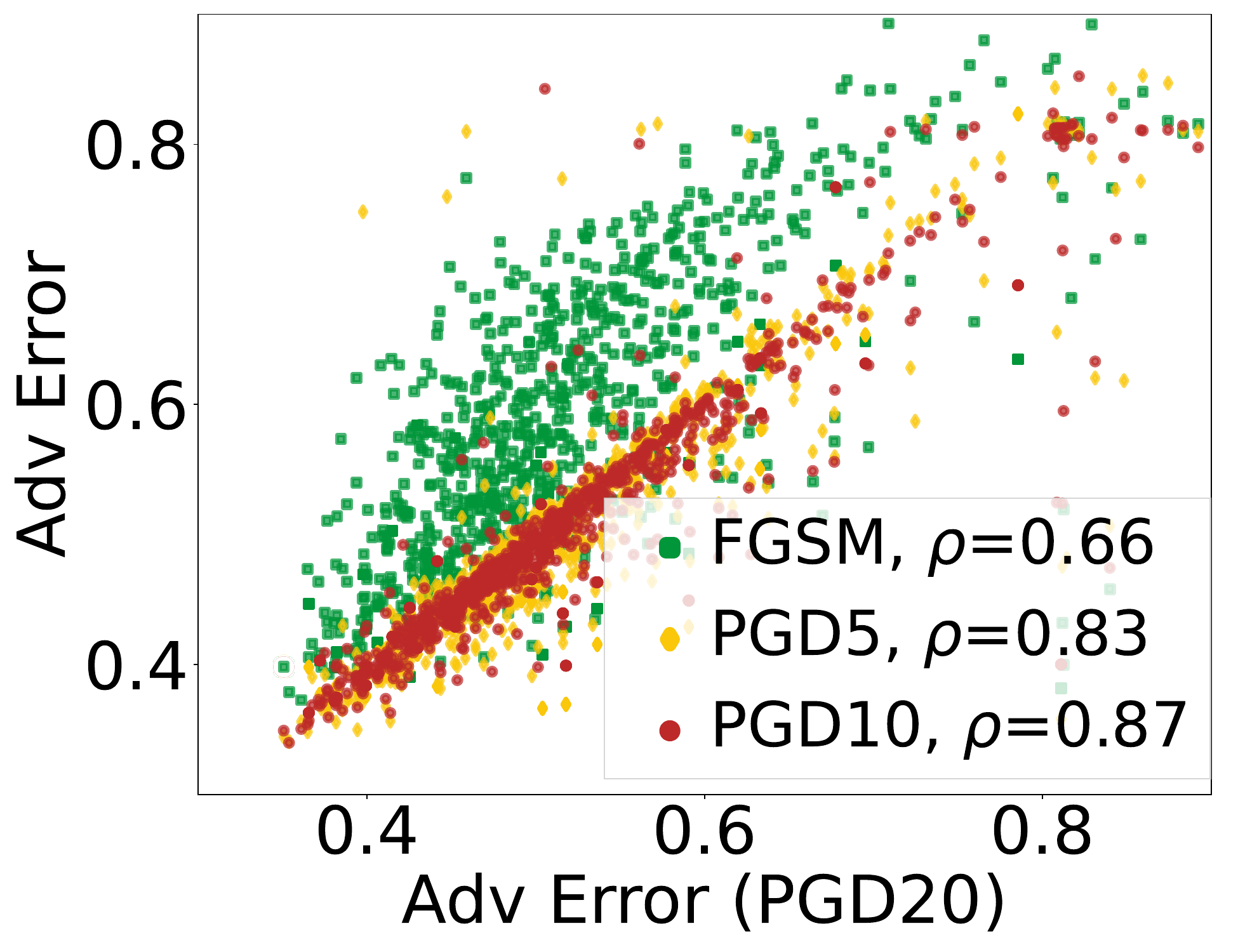}
        \caption{Adv. Error Corr. \\ResNet18/SVHN $\epsilon$\textdblhyphen4}
        \label{fig:correlation_advacc_svhn_4}
    \end{subfigure}  
    \begin{subfigure}[b]{0.24\textwidth}
        \includegraphics[ width=\textwidth]{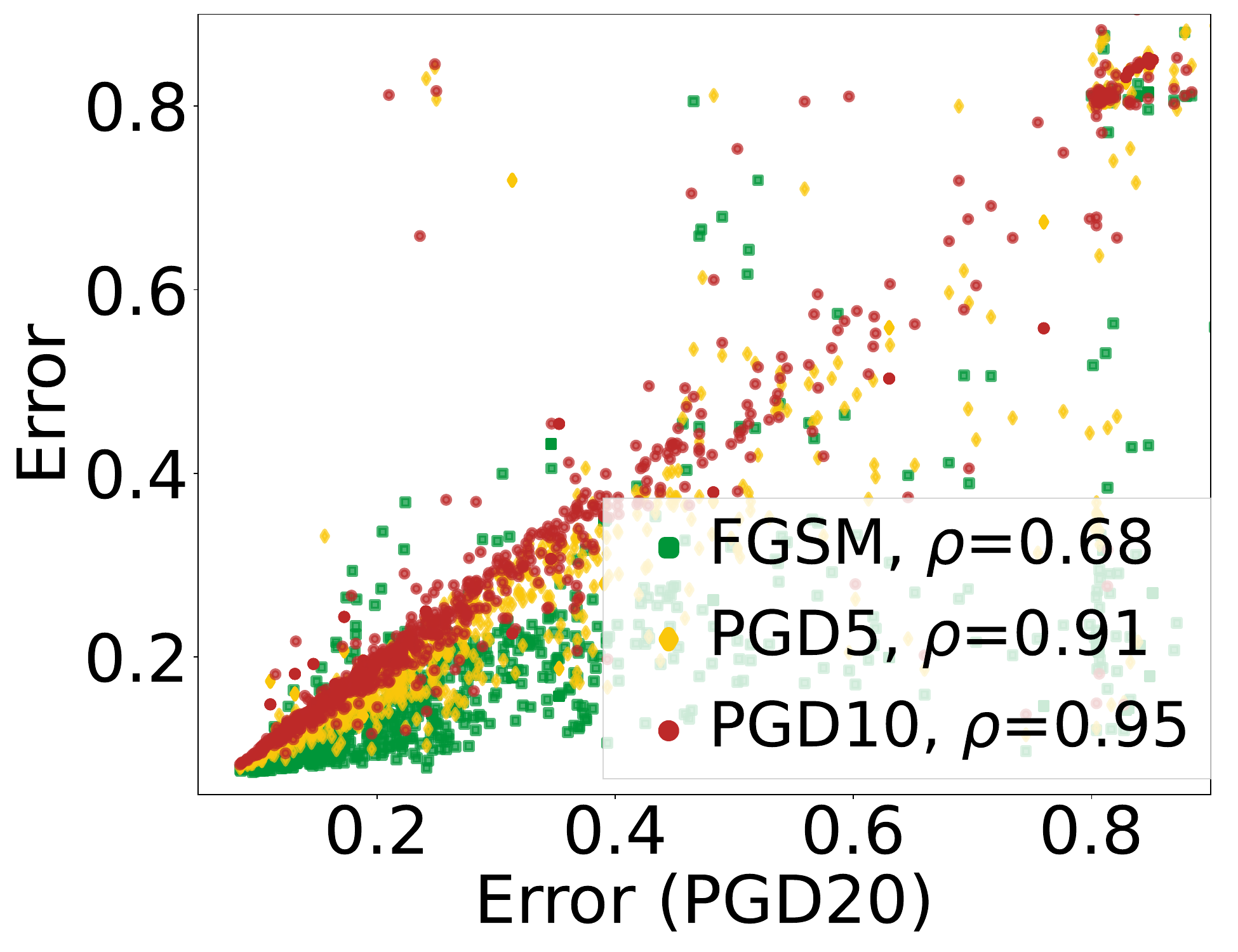}
        \caption{Error Corr. \\ResNet18/SVHN $\epsilon$\textdblhyphen8}
        \label{fig:correlation_acc_svhn_8}
    \end{subfigure}
    \begin{subfigure}[b]{0.24\textwidth}
        \includegraphics[ width=\textwidth]{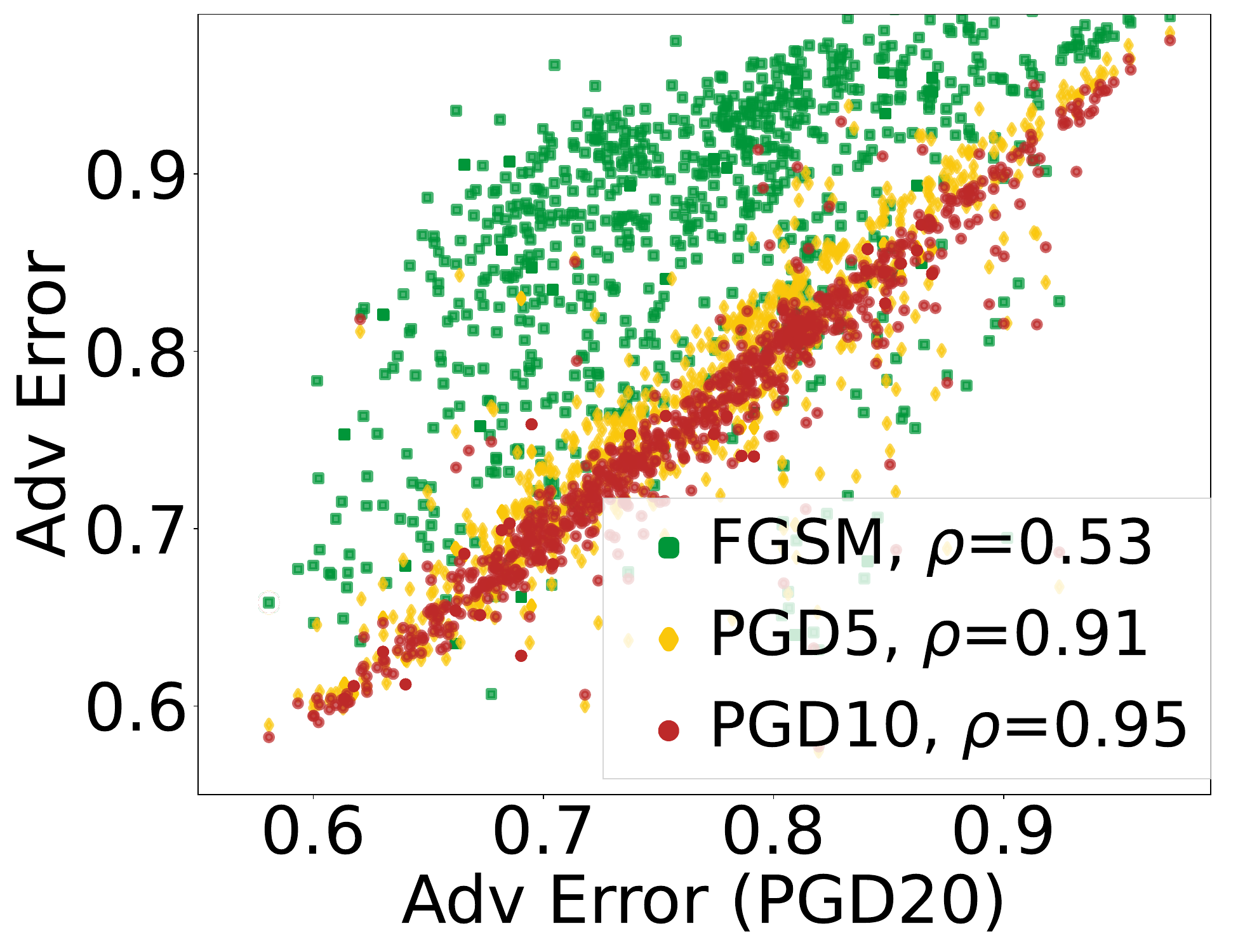}
        \caption{Adv. Error Corr. \\ResNet18/SVHN $\epsilon$\textdblhyphen8}
        \label{fig:correlation_advacc_svhn_8}
    \end{subfigure}
    
\vspace{3mm}
    \begin{subfigure}[b]{0.24\textwidth}
        \includegraphics[ width=\textwidth]{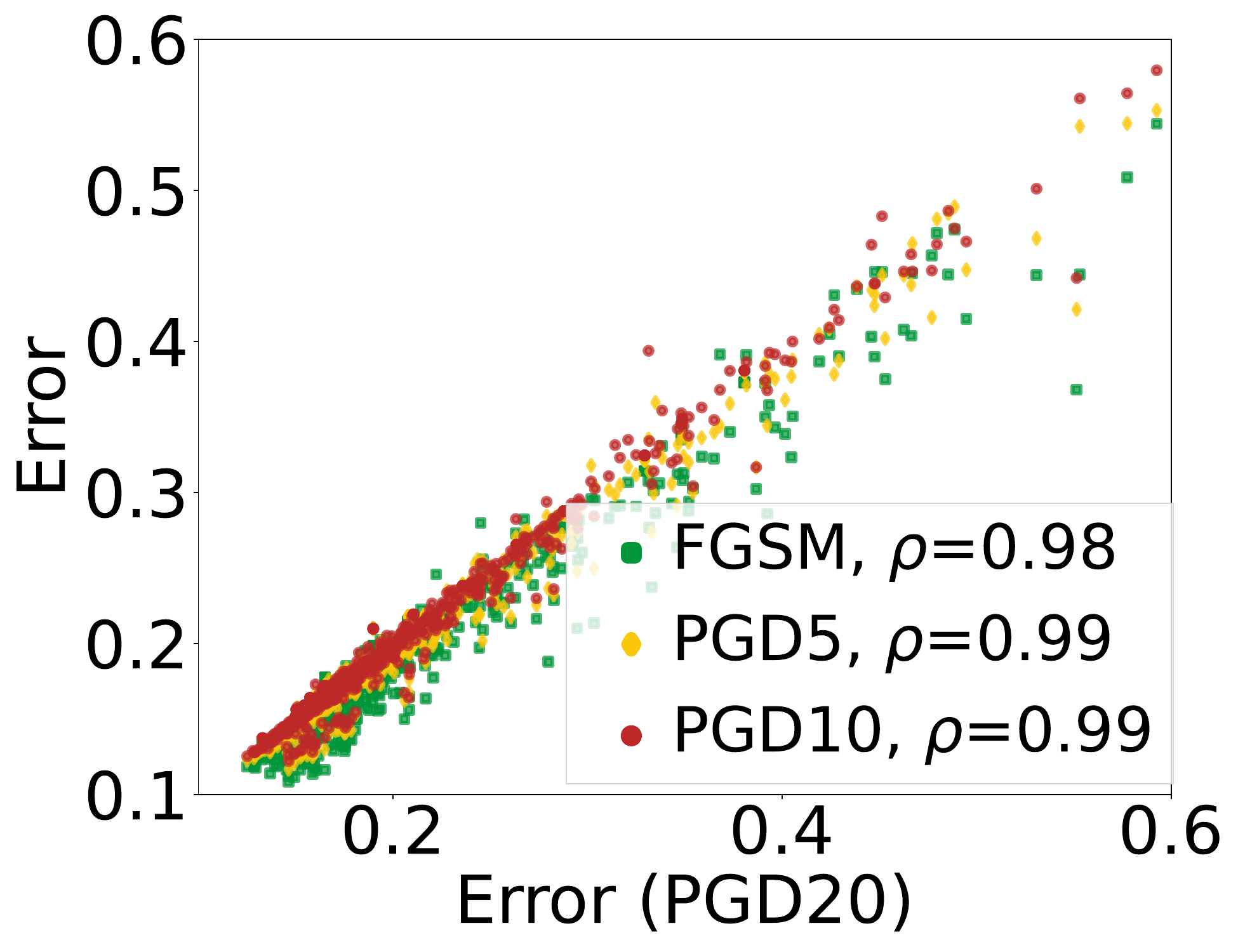}
        \caption{Error Corr. \\CNN/Cifar10 $\epsilon$\textdblhyphen8}
        \label{fig:correlation_acc_cifar10_8}
    \end{subfigure}
    \begin{subfigure}[b]{0.24\textwidth}
        \includegraphics[ width=\textwidth]{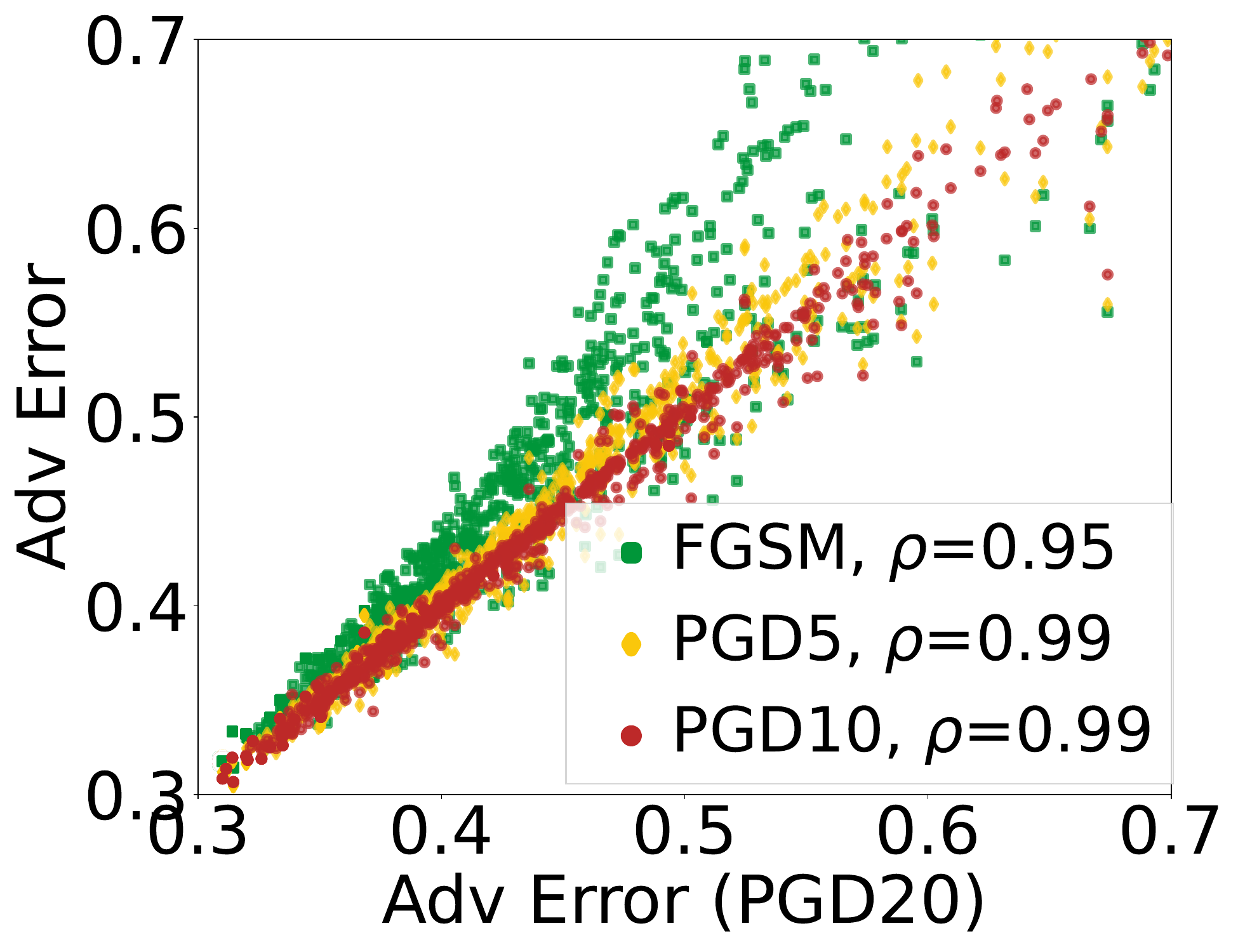}
        \caption{Adv. Error Corr. \\CNN/Cifar10 $\epsilon$\textdblhyphen8}
        \label{fig:correlation_advacc_cifar10_8}
    \end{subfigure}
    \begin{subfigure}[b]{0.24\textwidth}
        \includegraphics[ width=\textwidth]{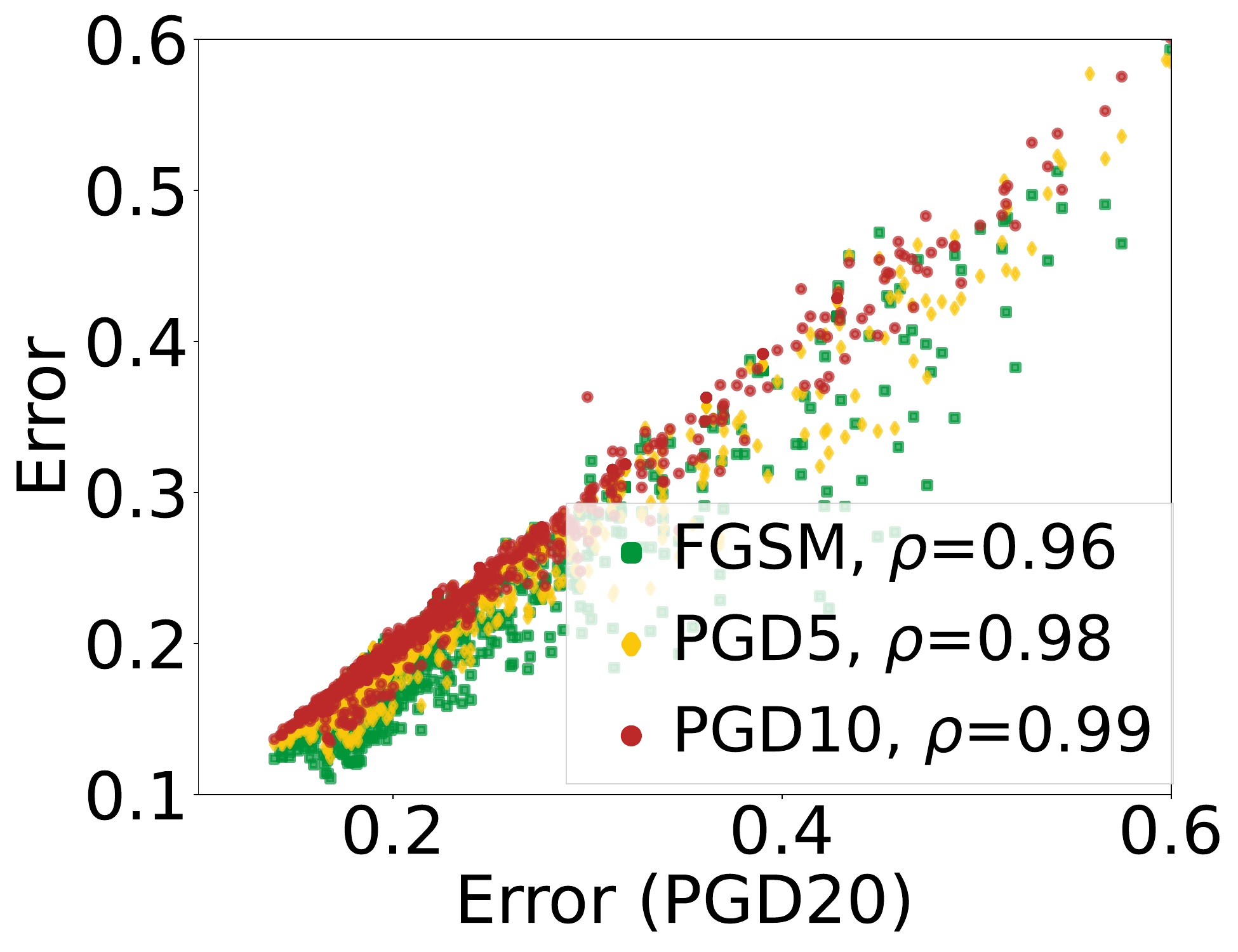}
        \caption{Error Corr. \\CNN/Cifar10 $\epsilon$\textdblhyphen12}
        \label{fig:correlation_acc_cifar10_12}
    \end{subfigure}
    \begin{subfigure}[b]{0.24\textwidth}
        \includegraphics[ width=\textwidth]{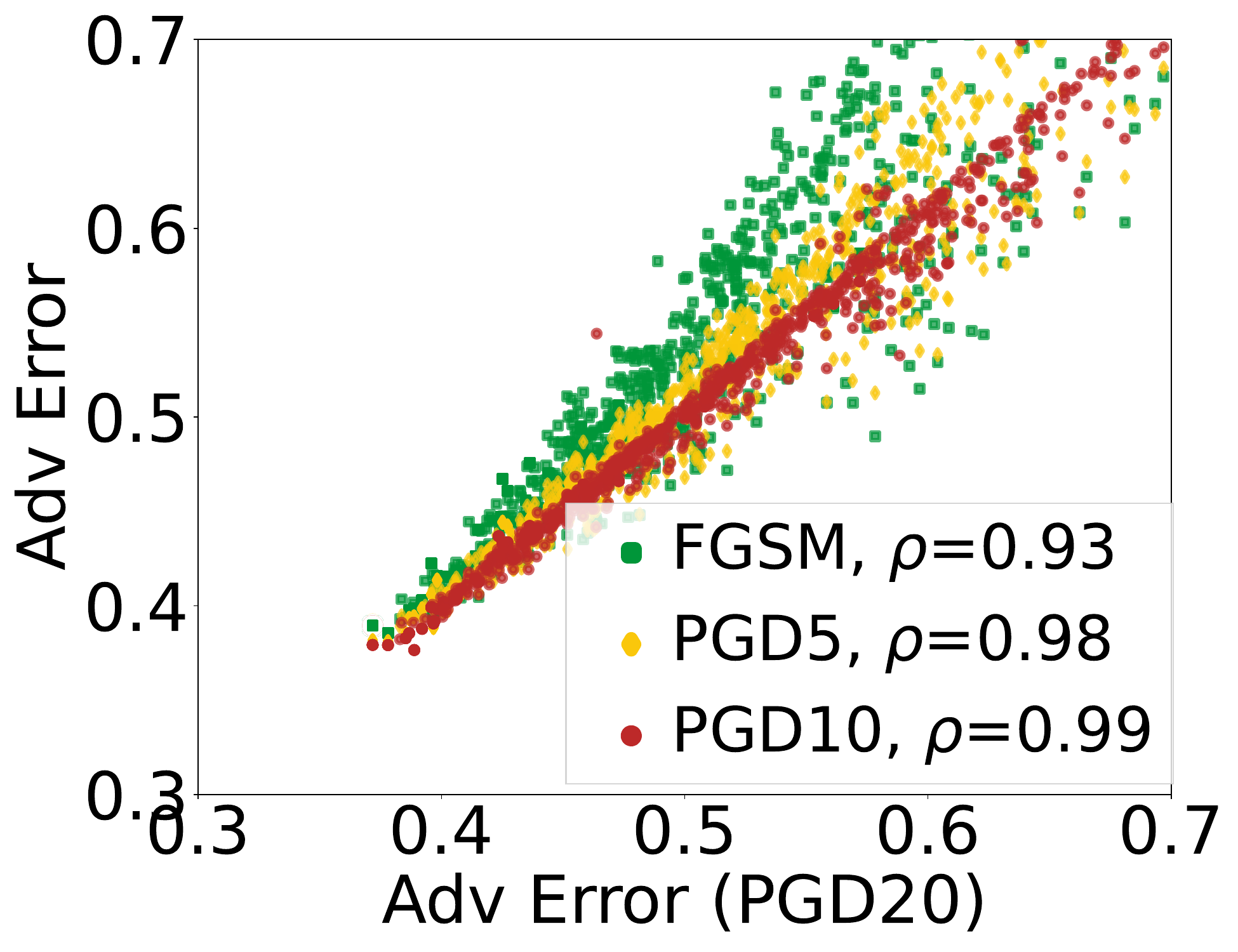}
        \caption{Adv. Error Corr. \\CNN/Cifar10 $\epsilon$\textdblhyphen12}
        \label{fig:correlation_advacc_cifar10_12}
    \end{subfigure}


    \captionsetup{justification=justified}
    \caption{Standard and adversarial error correlation  between PGD20 and FGSM, PGD5, and PGD10 varying the bound $\epsilon$ 
    and CDF of the training time reduction obtained using cheaper AT algorithms w.r.t PGD20 (Figure~\ref{fig:cost_imageNet}). 
    }
     \label{fig:correlation}

\end{figure*}

So far, we have shown that in adversarial settings the complexity of the HPT problem is exacerbated due to the need for optimizing a larger HP space. 
In this section, we show that, fortunately, AT provides also new opportunities to reduce the HPT cost. 
Specifically, we propose and evaluate a novel idea: leveraging alternative AT methods, which impose lower computational costs but provide weaker robustness guarantees to sample HP configurations in a cheap, yet informative, way. As discussed in Section~\ref{sec:3_related_word}, PGD is an iterative method, where each iteration refines the perturbation with the objective of maximizing loss. Hence, a straightforward way to reduce its cost (at the expense of robustness) is to reduce the number of executed iterations. We also note that the computational cost of FGSM is equivalent to that of a single PGD iteration.

We build on these observations to propose incorporating the number of PGD iterations as an additional fidelity dimension in multi-fidelity HPT optimizers, such as taKG~\cite{takg}. 
We choose to test the proposed idea with taKG since this technique supports the use of an arbitrary number of fidelity dimensions (e.g., dataset size and number of epochs) and determines how to explore the multi-dimensional fidelity via black-box modeling techniques (see Section~\ref{subsec:HPT}). 

In order to assess the soundness and limitations of the proposed approach, we first analyze the correlation of the standard and adversarial error between HP configurations that use PGD with 20 iterations (which we consider as the maximum-fidelity/full budget) vs.~PGD with 10 and 5 iterations and FGSM (which, as already mentioned, is computationally equivalent to 1 iteration of PGD). In Figure~\ref{fig:correlation},  we observe that the correlation varies for different bounds on adversarial perturbations across the considered models/datasets. 
We omit the correlation for ResNet50/ImageNet using $\epsilon$=4 since the results are very similar to $\epsilon$=2.
The scatter plots  clearly show the existence of a very strong correlation (above 95\%) for all the considered methods for ResNet50/ImageNet and for all the considered bounds. For ResNet18/SVHN and CNN/Cifar10, the correlation of PGD 5 and 10 iterations remains quite strong (always above 80\% and typically above 90\%), whereas lower correlations (as low as 53\%) can be observed for FGSM, especially when considering adversarial error and larger values of $\epsilon$ . This is expected, as previous works~\cite{fgsm2,catastrophic_overfitting} had indeed observed that FGSM tends to be less robust than PGD when larger $\epsilon$ values are used (being subject to issues such as catastrophic-overfitting that lead to a sudden drop of adversarial accuracy). Still, even for FGSM, the correlation is always above 90\% with CNN/Cifar10 and is relatively high (around 70\%) also with ResNet18/SVHN for the smaller considered bound ($\epsilon=4$).

We also report, in Figure~\ref{fig:cost_imageNet}, 
 the CDF of the training time reduction using FGSM, PGD with 5 and 10 iterations, and ST w.r.t.~PGD20 for ResNet50/ImageNet. 
The CDFs show that the training time reductions for a given AT method vary since the ratio of computed adversarial examples depends on the \%RAT and \%AE parameters. Overall, the maximum (median) training time reduction is approximately 
83\% (54\%), 66\% (42\%), 47\% (28\%), and 86\% (53\%)
for FGSM, PGD5, PGD10, and STD, compared to PGD20, which confirms that leveraging these ``cheap'' surrogate methods can significantly reduce the cost of testing HP configurations.

\begin{figure*}[t]
\centering
    \begin{subfigure}[b]{0.32\textwidth}
        \includegraphics[ width=\textwidth]{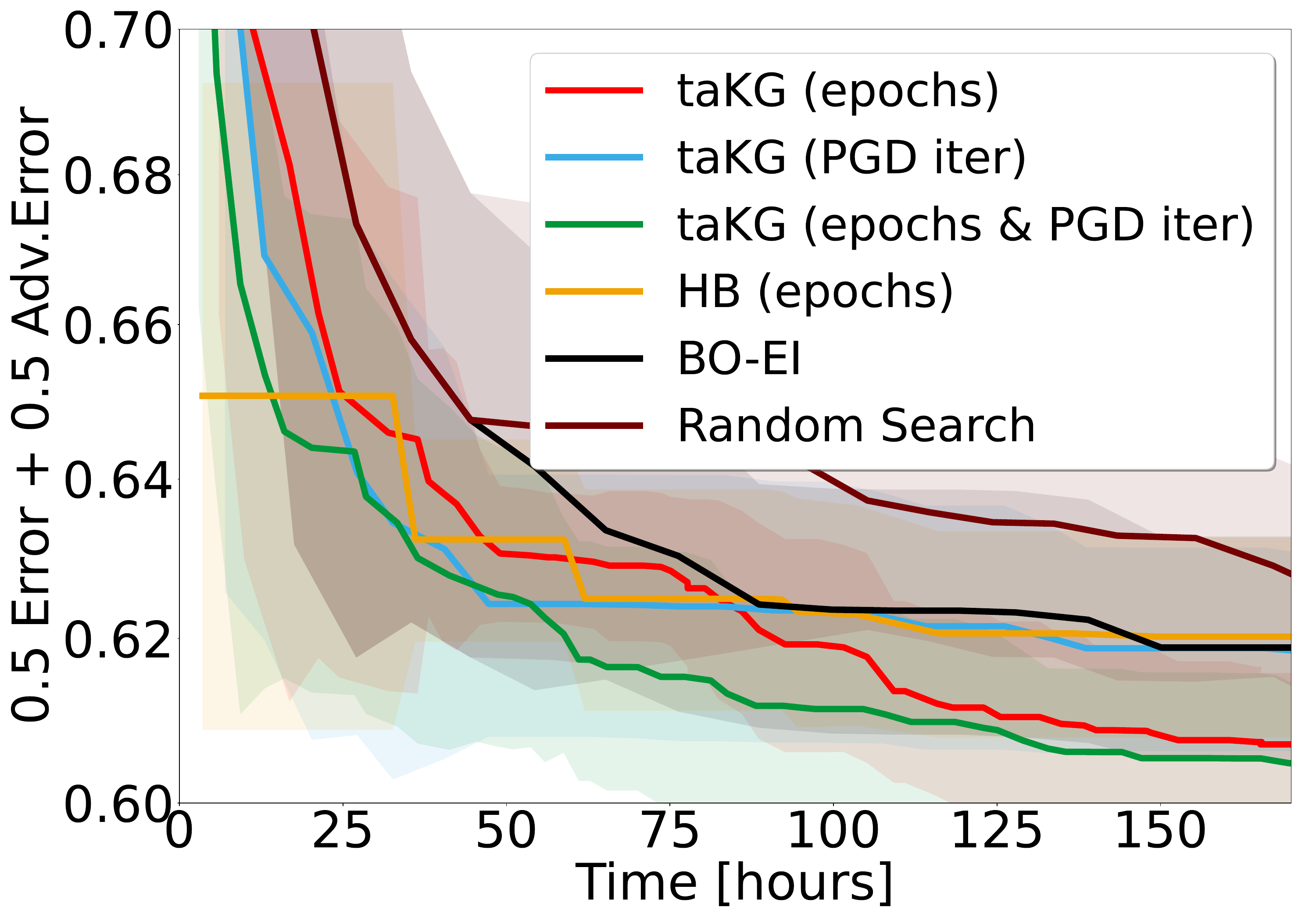}
        \caption{ResNet50/ImageNet $\epsilon$\textdblhyphen2}
        \label{fig:opt_imageNet}
    \end{subfigure}
    \begin{subfigure}[b]{0.32\textwidth}
        \includegraphics[ width=\textwidth]{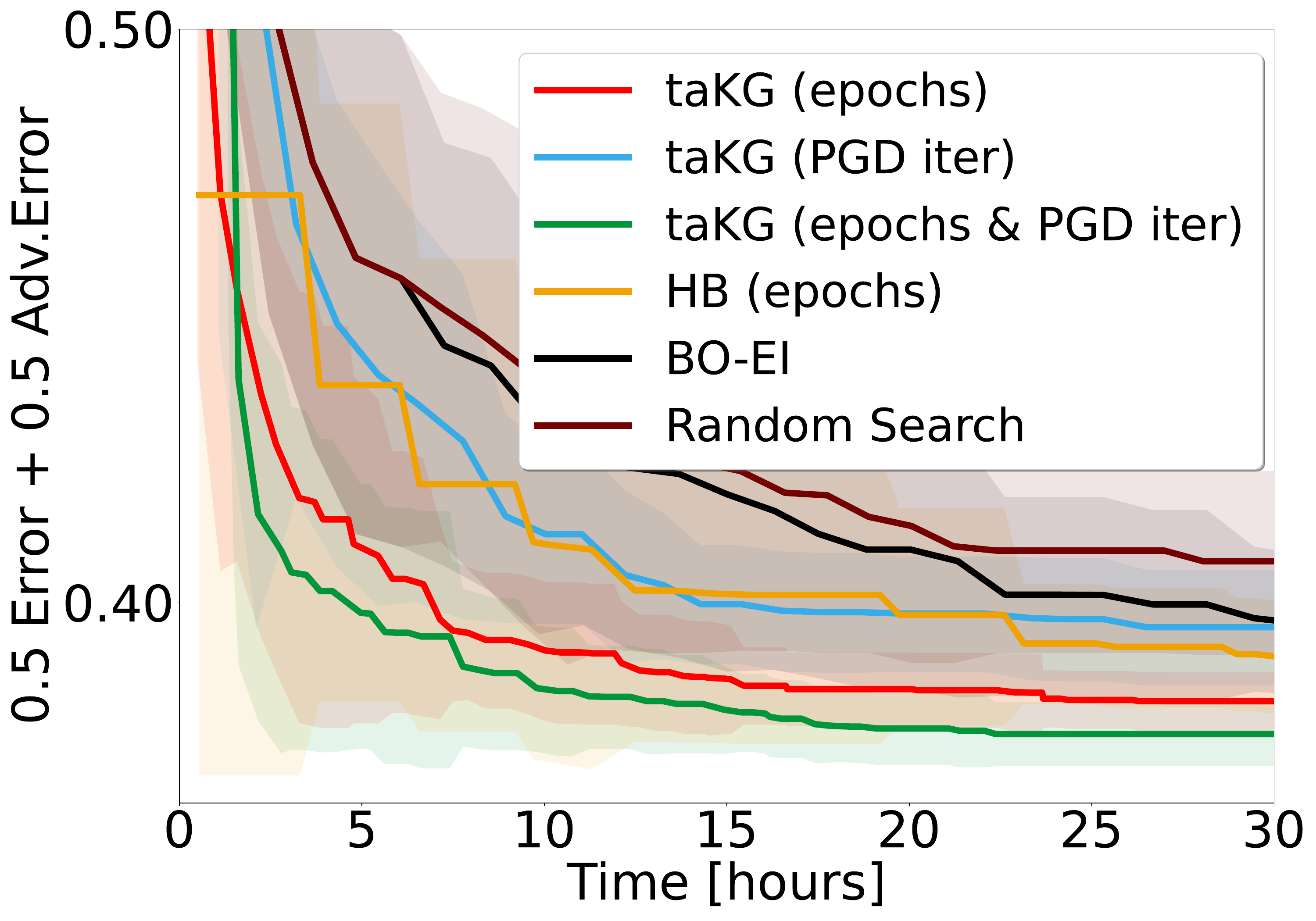}
        \caption{ResNet18/SVHN $\epsilon$\textdblhyphen8}
        \label{fig:opt_svhn}
    \end{subfigure}
    \begin{subfigure}[b]{0.32\textwidth}
        \includegraphics[width=\textwidth]{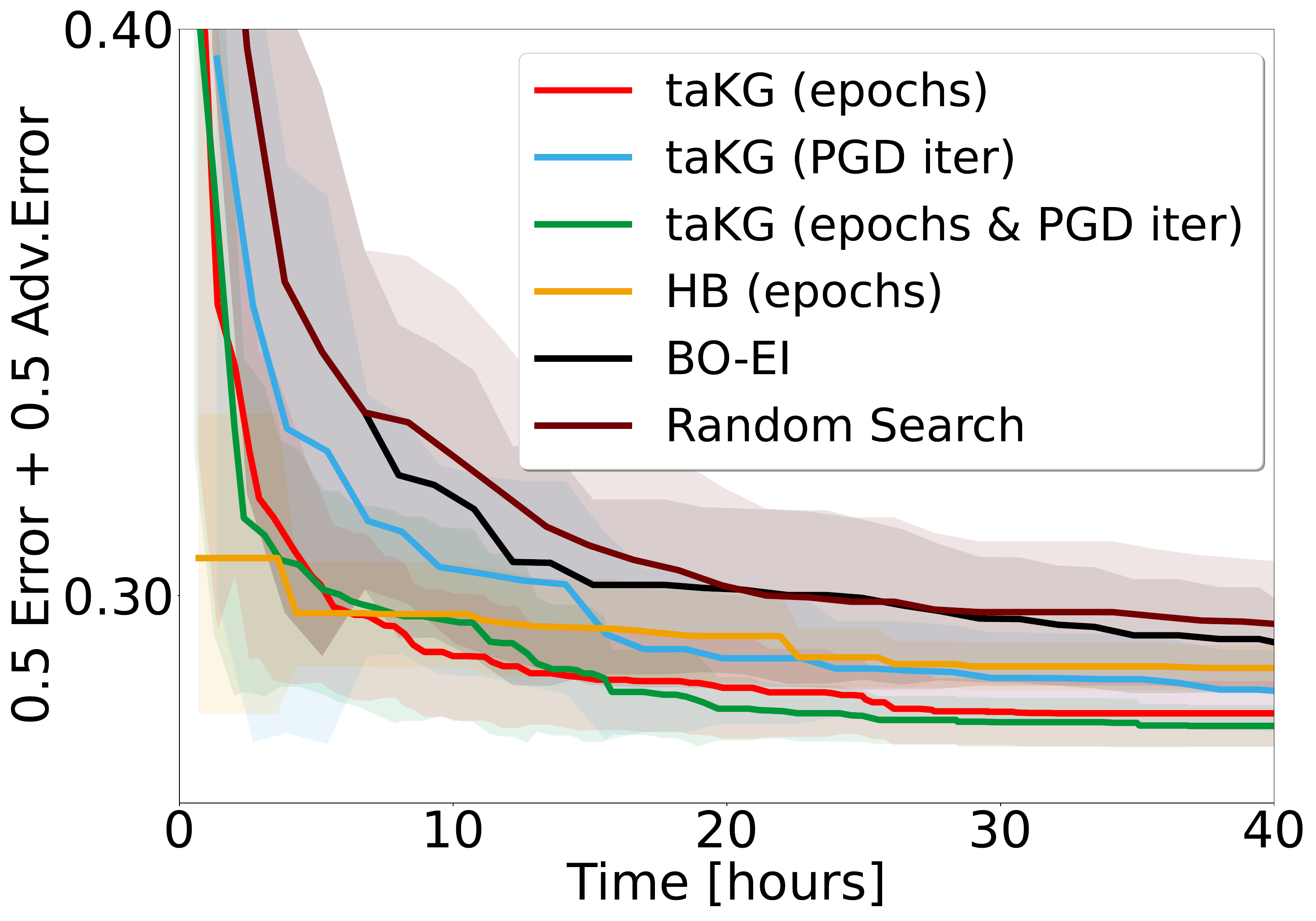}
        \caption{CNN/Cifar10 $\epsilon$\textdblhyphen12}
        \label{fig:opt_cifar10}
    \end{subfigure}
    
    \caption{Average standard and adversarial error using different optimizers.}
     \label{fig:optimizers}
\end{figure*}

Supported by these findings, we evaluate our proposal by integrating the number of PGD iterations as an additional fidelity dimension in taKG~\cite{takg}. As discussed in  Section~\ref{sec:3_related_word}, taKG is a HPT method that natively supports the use of multiple fidelity types, which we refer to as fidelity dimensions, e.g., number of epochs, input size, and dataset size. 
Based on the results of the previous section, we independently optimize the HPs of the ST and AT phases, which yields a search space composed of a total of nine HPs (Table~\ref{tab:fixed_hp}). 
The following multi-fidelity solutions are compared:

\begin{itemize}
\item \textit{taKG (epochs, PGD iter)}: the proposed solution, which uses taKG as the underlying HPT method and employs as fidelity the number of epochs and the number of PGD iterations. We discretize these 2 dimensions (Table~\ref{tab:budgets}). 
\item \textit{taKG (epochs)}: taKG  using as fidelity only the number of epochs. 
\item \textit{taKG (PGD iter)}: taKG using  as fidelity only the number of PGD iterations.
\item \textit{HB (epochs)}: HyperBand~\cite{hyperband}, a popular (single-dimensional) multi-fidelity optimizer, which uses  the number of epochs as fidelity.
\end{itemize}

We further compare against BO using EI as acquisition function and Random Search (RS). These optimizers only perform high-fidelity evaluations.
The evaluation of these alternative solutions is performed by exploiting the dataset already described in Section~\ref{sec:experiment}, which specifies the model quality (error and adversarial error) for all possible HPs, $\epsilon$ and fidelity settings reported in Tables~\ref{tab:fixed_hp},~\ref{tab:bounds} and \ref{tab:budgets}.

We define the optimization problem as follows: 
\begin{equation}
    \underset{x}{\min} \text{ } \lambda \cdot \text{Error}(x,s=1) + (1 -\lambda) \cdot \text{Adv.Error}(x,s=1) 
\end{equation} 
where $x$ is a vector defining the HPs, $s$ is a vector that encodes the ratio of budget allocated for each fidelity dimension, and $\lambda$ is a weight factor that we set to 0.5 to equally balance the standard and adversarial errors. For a fair comparison, when a single fidelity dimension (e.g., epochs) is used, we set the other fidelity dimension (e.g., PGD iterations)  to its maximum value. 
%
We run each optimizer using 20 independent seeds. We set the bound $\epsilon$ to 2, 8, and 12 to optimize ResNet50/ ImageNet, ResNet18/SVHN, and CNN/Cifar10. 
Based on Figure~\ref{fig:correlation}, the three settings correspond to scenarios with relatively high, low, and medium correlations for the budget dimension defined by PGD iterations, respectively.

Figure~\ref{fig:optimizers} reports the average and standard deviation of the optimization goal (i.e., $0.5\cdot\text{Error} + 0.5\cdot\text{Adv.Error}$) as a function of the optimization time for the different HPT optimizers and different models/datasets.
The results show that the proposed solution, which adopts PGD iterations as extra fidelity along with the number of epochs (taKG - epochs \& PGD iter), clearly outperforms all the alternative solutions in ResNet50/ImageNet and ResNet18/SVHN. 
The largest gains can be observed with ResNet18/SVHN (Fig.~\ref{fig:opt_svhn}). Here, at the end of the optimization process, the proposed solution identifies a configuration of the same quality as the ones suggested by the other baselines, namely taKG - epochs, HB, BO-EI, by achieving speed-ups of 2.1$\times$, 3.7$\times$, and 5.4$\times$, respectively.
Using the same metric (time spent to recommend a configuration of the same quality as the ones suggested by the other optimizers at the end of the optimization process) with ResNet50/ImageNet, the proposed solutions achieve slightly smaller, but still significant speed-ups, namely 1.28$\times$, 1.97$\times$ and 2.45$\times$ w.r.t.~taKG - epochs, HB, BO-EI. Interestingly, with ResNet50/ImageNet the proposed solution provides solid speed-ups also during the first stage of the optimization. Specifically, if we analyze the first half of the optimization process  (corresponding to approx. 83 hours (see Figure~\ref{fig:opt_imageNet})) the proposed solution identifies configurations of the same quality as taKG - epochs, HB, BO-EI, with speed-ups of 1.7$\times$, 2$\times$ and 2.6$\times$, respectively.

With CNN/Cifar10 (Figure~\ref{fig:opt_cifar10}), the proposed approach remains the best-performing solution, although with smaller gains when compared to taKG with epochs. Still, the proposed solution can identify configurations with the same quality as the best alternative (taKG epochs) by saving approx.~40\% of the time (i.e., in 22 hours vs. 32 hours). We argue that the gains with CNN/Cifar10 are relatively lower than in the other scenarios considered in Figure~\ref{fig:optimizers} since those models are larger and more complex. As such, they benefit more from the cost reduction opportunities provided by using a reduced number of PGD iterations.

We also observe that the exclusive use of  PGD iterations with taKG yields worse performance than using solely number of epochs. This is not surprising, given that number of epochs is arguably one of the most direct ways of controlling the cost of configuration sampling and is, indeed, among the most commonly adopted budgets in multi-fidelity optimizers~\cite{fabolas,hyperband,hyperjump}. 
This result confirms that PGD iterations represent a valuable mean to accelerate multi-fidelity HPT optimizers to train robust models and that it complements, but does not replace, "conventional" budget settings like number of epochs or dataset size.

\section{Conclusions and Future work}
\label{sec:8_conclusion}

This paper focused on the problem of HPT for robust models. By means of an extensive experimental study, we first quantified  the relevance of  
independently tuning  the HPs used during standard and adversarial training.
We then proposed and evaluated a novel fidelity dimension that becomes available in the context of AT. Specifically, we have shown that cheaper AT methods can be used to obtain inexpensive estimations of the quality achievable via expensive state-of-the-art AT methods and that this information can be effectively exploited to accelerate HPT. 
We extended taKG, a state-of-the-art HPT method, by incorporating  the PGD iterations as an additional fidelity dimension (along with the number of epochs) and achieved cost reductions by up to 2.1$\times$.

It is worth noting that the idea of employing ``cheap'' AT methods as proxies to estimate the quality of HP configurations with more robust/expensive methods is generic, in the sense that it can be applied, at least theoretically, to any multi-fidelity optimizer. As part of our future work, we plan to integrate this novel approach in a new HPT framework, specifically designed to cope with adversarially robust models.

\begin{acknowledgments}
This work was supported by the Fundação para a Ciência e a Tecnología (Portuguese Foundation for Science and Technology) through the Carnegie Mellon Portugal Program under grant SFRH/BD/151470/2021 via projects with reference  
UIDB/50021/2020 and 
C645008882\mbox{-}00000055.PRR, 
by the NSA grant H98230-23-C-0274,
and by the Advanced Cyberinfrastructure Coordination Ecosystem: Services \& Support (ACCESS) program, where we used the Bridges-2 GPU and Ocean resources at the Pittsburgh Supercomputing Center through allocation CIS220073, which is supported by National Science Foundation grants \#2138259, \#2138286, \#2138307, \#2137603, and \#2138296.



\end{acknowledgments}

\bibliography{biblio}

\end{document}